%% file: main.tex
\documentclass{article}

\usepackage{arxiv,times}
\usepackage[utf8]{inputenc}
\usepackage[T1]{fontenc}
\input{math_commands.tex}

\usepackage{amsmath}
\usepackage{bm}
\usepackage{algorithm}
\usepackage{algpseudocode}
\usepackage[dvipsnames]{xcolor}
\usepackage{mathtools}
\usepackage{booktabs}
\usepackage{amssymb}
\usepackage{amsthm}
\usepackage{hyperref}
\usepackage{url}
\usepackage{graphicx}
\usepackage{xspace}
\usepackage{adjustbox}
\usepackage[font=small,labelfont=bf]{caption}
\usepackage{natbib}
\usepackage{tabularx}
\usepackage{array}
\usepackage{colortbl}
\usepackage{multirow}
\usepackage{enumitem}
\usepackage{pifont}
\usepackage{listings}
\usepackage{tikz}
\usepackage{adjustbox}
\usetikzlibrary{positioning,arrows.meta,fit,backgrounds,calc,shadows,shapes.geometric}

\usepackage{wrapfig}

\theoremstyle{remark}

\providecommand{\method}{\textit{DAiSy}\xspace}

\providecommand{\val}{\gV}
\providecommand{\vphi}{\bm{\phi}}

\providecommand{\vdelta}{\bm{\delta}}

\providecommand{\tphi}{\tilde{\bm{\phi}}}

\definecolor{revcolor}{HTML}{D00000}

\definecolor{daisyrow}{RGB}{255,246,204}
\definecolor{darkdaisyrow}{RGB}{255,232,208}
\definecolor{DarkRed}{rgb}{0.5,0.1,0.1}
\definecolor{DarkBlue}{rgb}{0.1,0.1,0.7}
\definecolor{myblue}{rgb}{0.00, 0.45, 0.85}
\definecolor{myred}{rgb}{0.84, 0.25, 0.00}
\definecolor{mygreen}{rgb}{0.10, 0.50, 0.20}
\hypersetup{colorlinks=true, linkcolor=DarkBlue, citecolor=DarkRed, urlcolor=DarkBlue,
  pdftitle={\textit{DAiSy}: Online Data Selection by Adaptive Utility Estimation for LLM Pretraining}}

\lstdefinestyle{py}{language=Python, basicstyle=\ttfamily\scriptsize, keywordstyle=\color{DarkBlue}\bfseries,
  commentstyle=\color{gray}\itshape, stringstyle=\color{DarkRed}, showstringspaces=false, columns=fullflexible,
  frame=single, framerule=0.3pt, rulecolor=\color{gray!60}, xleftmargin=2pt, xrightmargin=2pt, aboveskip=4pt, belowskip=4pt,
  breaklines=true}
\newcommand\mapemoji{\raisebox{-2pt}{\includegraphics[width=1em]{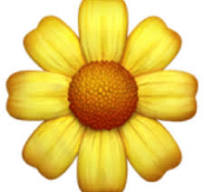}}}

\title{\textit{DAiSy\,}\mapemoji{}: Online Data Selection by Adaptive Utility Estimation for LLM Pretraining}
\author{Simin Fan, Martin Jaggi\\ EPFL\\ \texttt{firstname.lastname@epfl.ch}}

\begin{document}
\maketitle

\begin{abstract}
Model-aware online data selection prioritizes the training documents that contribute the most given the model’s current state.
Existing online selectors rank candidates upon gradient-based data utility scores, which requires the full forward-backward pass of the main model on every candidate document. Scoring each document therefore pays as much as training on it, yet only a small fraction is used for the real model update.
We introduce \method, a lightweight linear auxiliary model that learns to predict data utility scores from frozen document embeddings, which eliminates the computation overhead and infrastructure burdens. 
The linear estimator is updated alongside the main language model, fitting to the exact gradient-based utility scores obtained from the backward pass of regular training steps. Consequently, scoring a candidate requires only a linear transformation, bypassing the costly gradient-alignment computation. 
We demonstrate its effectiveness and efficiency for LLM pretraining in English and multilingual scenarios, with various steering target designs. The improvements from \method can consistently survive the instruction-tuning phase. We release our codebase for reproducibility \url{https://github.com/Olivia-fsm/daisy_official}.
\end{abstract}
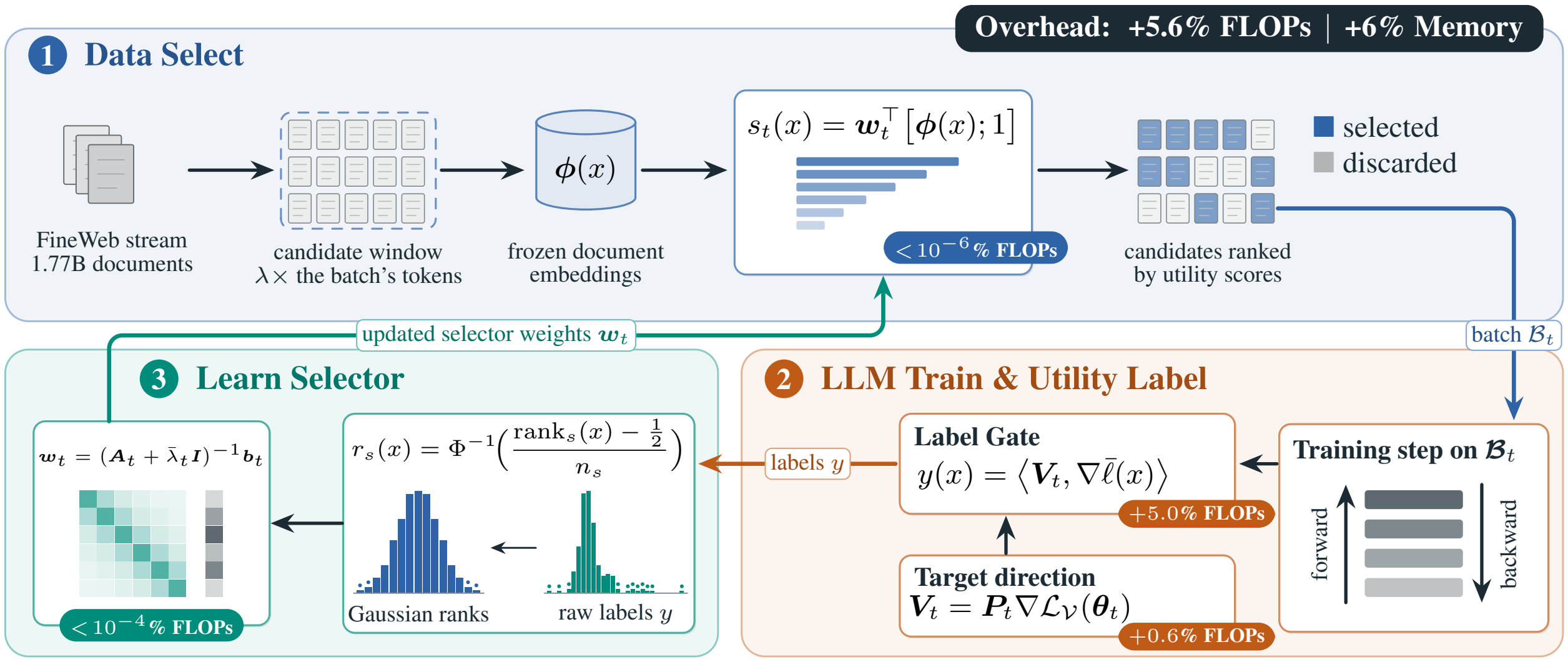
\begin{figure}[ht!]
\centering
\input{figures/pipeline_diagram}
\caption{\textbf{The \method pipeline inside a training run.} (1) \textbf{\textit{Data Select}}: Linear selector processes the candidate pool. No pass through the LLM; (2) \textbf{\textit{Train and Label}}: LLM's standard backward pass holds each trained document’s exact gradient utility; (3) \textbf{\textit{Learn Selector}}: An online ridge regression maps embeddings to those labels, so the linear selector keeps up with the model. }
\label{fig:pipeline}
\end{figure}
\tableofcontents
\newpage
\clearpage
\section{Introduction}
Training data is one of the strongest levers that determine the quality of language model pretraining~\citep{albalak2024survey,penedo2024fineweb,li2024datacomp}. 
Static selectors, such as quality classifiers and target-matching resampling, curate the documents once before training and are therefore not adaptive to evolving model states~\citep{wettig2024qurating,penedo2024fineweb,xie2023dsir,grangier2025crisp}. 
To keep track of the model's stage-wise data preference, online \emph{model-aware} selectors instead estimate how much a candidate would help the model at its current learning status~\citep{mindermann2022prioritized,engstrom2024dsdm,yu2024mates,wang2024greats}, which adapts to the model's training dynamics during training. \citet{fan2026optiselect} further derive the optimal oversampling ratio for online data selection under the trade-off between efficiency and performance.
However, the severe computation and memory overhead incurred by existing online data selection pipeline limits its scalability in production-level model training. \citet{wang2024greats} and \citet{wang2026opus} apply an \textit{oversampling}-then-\textit{selection} paradigm: at each training step, a larger batch with $2\times$ candidates is sampled and scored with gradient-based utilities, and then only half of the batch is selected and applied in model updates. The overhead comes from two sources: First, gradient-based utility scoring takes the full forward-backward pass on every candidate in the oversampled batch, where data selection already consumes $2\times$ more FLOPs than the training cost; Second, it requires loading $2\times$ more data into the GPU and RAM, which is a severe waste of hardware bandwidth. 

We propose \method (online \underline{\textbf{DA}}ta \underline{\textbf{S}}election by adaptive utilit\underline{\textbf{Y}} estimation) as a lightweight but effective online selection pipeline for LLM pretraining. 
Inspired by the expressivity of a linear predictor on data utility \citep{ilyas2022datamodels}, we train a linear network alongside the primary LLM that learns to predict the gradient-based data utility scores from frozen document embeddings. 
Following the ghost-engine technique proposed by \citet{wang2024greats}, one standard forward-backward optimization step yields the first-order gradient-based utility scores for all the documents in the training batch as a by-product with very few extra FLOPs, \method therefore eliminates the computation overhead and infrastructure burdens. Figure \ref{fig:pipeline} shows the complete online data selection pipeline with \method: At each training step, the linear data selector first estimates the utility scores on an oversampled batch of documents and selects the top-ranked documents to the LLM; then the LLM performs the standard training step on the selected batch, while providing the exact gradient-based utility scores. The Gaussianized ranks of the exact utilities then serve as ground-truth labels for the linear data selector; next, the linear data selector fits the updated utility labels via ridge regression and is updated for the next iteration. With \method, the primary LLM only performs standard training steps without extra input data or extra forward-backward passes, while the online data selector stays synchronized as the ground-truth utility scores are computed with the latest model states. 
\paragraph{Contributions.}
\begin{itemize}[leftmargin=1.4em,itemsep=1pt,topsep=2pt]
\item We propose \method (\S\ref{sec:method}), an effective and scalable online data selection pipeline for LLM training. The lightweight and model-aware selector significantly improves data efficiency in LLM pretraining without incurring significant computation or memory overhead. 
\item We empirically evaluate \method in pretraining runs at various model scales (\S\ref{sec:results}), which significantly outperforms uniform sampling, static embedding similarity baseline and prior online data selection baselines (GREATS \citep{wang2024greats}, OPUS \citep{wang2026opus} and MATES \citep{yu2024mates}). We also show that the pretraining performance gain earned from our data selection survives the instruction tuning stage (\S\ref{sec:sft}).
\item We demonstrate the effectiveness of \method across various setups and target designs, including task-oriented pretraining (\S\ref{sec:benchmark-target}), general web-targeted pretraining (\S\ref{sec:webtargets}), multilingual pretraining (\S\ref{sec:multilingual}) and a token-limited scenario (\S\ref{sec:budget}). \method effectively identifies useful data with consistent downstream improvements, while the static \textit{Similarity} baseline is vulnerable under heterogeneous multilinguality and the overfitting risk in token-limits.
\end{itemize}

\section{Related Work}
\label{sec:related}
\textbf{Data attribution.} Influence functions~\citep{hampel1974influence,koh2017understanding,grosse2023studying} and trajectory-based scores~\citep{pruthi2020estimating,wang2025datashapley} measure the effect of a training example on a target loss through inner products of gradients. Random projections~\citep{park2023trak,xia2024less} and inner products computed inside a single backward pass~\citep{wang2024greats,wang2025datashapley} make these scores affordable at scale, and datamodels regress training outcomes on the membership of examples in the training set~\citep{ilyas2022datamodels,engstrom2024dsdm}. \method computes the same first-order score, but only for the documents that training has already processed, and uses it to supervise a selector rather than to select directly.

\textbf{Static and mixture-level selection.} Static methods score each document once, before training. Filters that rely on heuristics, classifiers, perplexity or the judgments of a language model keep the documents of high quality~\citep{penedo2024fineweb,li2024datacomp,marion2023less,sachdeva2024train,wettig2024qurating}, and importance resampling moves the corpus towards a target distribution~\citep{xie2023dsir}. Mixture-level methods reweight whole domains, in some cases during training~\citep{xie2023doremi,fan2024doge,albalak2023odm}, but do not distinguish the documents within a domain. Our static baselines, embedding similarity and DSIR, are of the first kind.

\textbf{Online selection.} Online methods choose the training data with a signal from the current model: the loss or the gradient norm of a candidate~\citep{loshchilov2016online,katharopoulos2018not,jiang2019accelerating}, a coreset that matches the full gradient~\citep{mirzasoleiman2020coresets,killamsetty2021gradmatch}, the reducible loss~\citep{mindermann2022prioritized} or, for language model pretraining, the alignment of the candidate's gradient with a target gradient~\citep{wang2024greats,fan2026optiselect,wang2026opus}. All of them pass every candidate through the model, so their cost grows with the number of candidates. \citet{kaddour2023notrain} find that such gains largely vanish for Transformer language models once this cost is charged, and we therefore compare at equal GPU-hours as well as at equal tokens. MATES~\citep{yu2024mates} amortizes the scoring with an influence model that is refitted at intervals, but it obtains its labels from separate model updates on probe documents and scores the pool with a second neural network after every refit. \method avoids both costs: its labels come from the backward pass of training, and a candidate is scored by a dot product with a precomputed embedding.

\textbf{Learned data valuation.} Closest to our approach are methods that learn a value estimator during training: meta-learned reweighting weights each example by the alignment of its gradient with a validation gradient~\citep{ren2018learning} or learns a small weighting network alongside the model~\citep{shu2019metaweightnet}, and DVRL trains a data-value estimator jointly with the predictor by reinforcement learning~\citep{yoon2020dvrl}. We did not implement those baselines because of the intractable cost from meta-learning on LLMs. In contrast, the estimator of \method is linear in a frozen embedding, its labels are per-document first-order utilities rather than a scalar reward, and applying it to a candidate requires no pass through the model. Its form is that of factorized models with cached item embeddings in recommendation and retrieval~\citep{koren2009matrix,covington2016youtube,yi2019sampling,johnson2019billion} and of recent decision models that avoid generation~\citep{kwok2026contrastivelanguagemodels,almeida2026jev}.

\section{\method: Online Data Selection with Adaptive Utility Estimation}
\label{sec:method}
\paragraph{Setup and Preliminaries.} At step $t$ the model has parameters $\vtheta_t$, learning rate $\eta_t$ and optimizer preconditioner $\mP_t$. Specifically, we apply AdamW~\citep{loshchilov2019adamw} optimizer throughout our experiments with $\mP_t = \mathrm{diag}(1/(\sqrt{\hat\vv_t}+\epsilon))$. 
Given a validation set $\val$ from a target distribution, we define the target loss as $\gL_\val$. To first-order approximation, an optimization step on document $x$ changes the target loss by
\begin{equation}
-\Delta\gL_\val\;\approx\;\eta_t\big\langle \mP_t\nabla\gL_\val(\vtheta_t),\ \nabla\bar\ell(x;\vtheta_t)\big\rangle ,
\label{eq:utility}
\end{equation}
where $\bar\ell(x)$ is the document's mean token loss.
During pretraining, each LLM optimization step gives exact first-order gradient-based utility scores following Eq.~\ref{eq:utility}. Meanwhile, the online data selector is trained to rank the documents along their utility scores. We walk through the \method pipeline in Figure~\ref{fig:pipeline} and provide detailed demonstrations on each component in this section.  
\subsection{Synchronized utility labels from the training backward pass}
\label{sec:labels}
Firstly, we explain how the utility scores are read off from the training backward pass using the ghost engine \citep{wang2024greats}.
Let $\mV_t=\mP_t\nabla\gL_\val(\vtheta_t)$ be the preconditioned target direction. For one linear layer with per-token input $\va$ and output gradient $\vdelta$, the parameter gradient of document $x$ is $\sum_{\text{tok}\in x}\vdelta\va^\top$, so this layer's contribution of data utility is computed as 
\begin{equation}
y_{\text{layer}}(x)=\big\langle\mV_{t,\text{layer}},\nabla\bar\ell(x)\big\rangle=\frac{1}{|x|}\sum_{\text{tok}\in x}\vdelta^\top\mV_{t,\text{layer}}\,\va .
\label{eq:ghost}
\end{equation}
The ghost engine inserts a hook on every linear layer which computes the layer-wise gradient-alignments (\eqref{eq:ghost}) during backward pass, sums it over each document's tokens, then passes the output gradient through the computation graph~\citep{wang2024greats}. 
Summed over all layers, it gives the document's utility score $y(x)$ in \eqref{eq:utility}. 
In our implementation, each training batch packs documents into rows with document-masked attention, therefore we can compute the exact document-wise losses and utilities. 
The target direction $\mV_t$ is refreshed every $R$ steps, which takes one forward and backward pass over the target documents, while the latest $\mV_t$ is adopted between refreshes. 

\paragraph{Computation Overhead.} Since scoring requires no extra forward or backward passes, the overhead comes from the vector inner products in \eqref{eq:ghost} which costs $5.0\%$ of training FLOPs, and the target direction $\mV_t$ refresh which consumes $\approx 0.6\%$ training FLOPs amortized by $R=200$ steps. Since ridge regression uses a closed-form solution and the linear inference upon document embeddings takes negligible ($<10^{-4}\%$) FLOPs, \method incurs a total overhead of $\approx 5.6\%$ FLOPs. 
Appendix~\ref{sec:costmodel} reports the measured overhead in FLOPs, GPU time and memory and compares it with the other baselines.
\begin{algorithm}[t!]
\caption{Training with \method}
\label{alg:companion}
\begin{algorithmic}[1]
\Require \small Language model $\vtheta$, documents with frozen embeddings $\tphi(x)$; train/target loss $\gL$ and $\gL_\val$; window factor $\lambda$; warmup step to start selector training $t_0$; label interval $L$; target refresh interval $R$; refit interval $U$; decay $\gamma$; mini-batch tokens $|\gB|$
\vspace{0.2em}
\State $\mA\leftarrow\vzero$,\quad $\vb\leftarrow\vzero$,\quad $\vw\leftarrow\text{none}$ 
\vspace{0.1em}
\For{step $t=0,1,2,\dots$}
\vspace{0.2em}
  \If{$t\ge t_0$ \textbf{and} $t\bmod R=0$}
    \State $\mV\leftarrow\mP_t\nabla\gL_\val(\vtheta_t)$ \Comment{refresh the target direction}
  \EndIf
    \If{$\vw=\text{none}$}
      \State $\gB\leftarrow$ uniform draw from corpus
    \Else
      \State $C\leftarrow$ next candidates window, holding $\lambda\times |\gB|$ tokens
      \State $\vs_t\leftarrow$ estimated utility scores $\vw^\top\tphi(x), \forall x\in C$ \Comment{\textit{Scoring} (Eq.~\ref{eq:score})}
      \State $\gB\leftarrow$ with the highest scores $\vs_t$ 
    \EndIf
    \vspace{0.1em}
    \State Model forward and backward on $\gB$
    \vspace{0.1em}
    \If{$\big(t\ge t_0$ \textbf{and} $t\bmod L=0\big)$}
      \State Collect $y(x)\leftarrow\langle\mV,\nabla\bar\ell(x)\rangle$ for every $x\in\gB$ \Comment{\textit{Train and Label} (Eq.~\ref{eq:ghost})}
    \EndIf
    \vspace{0.1em}
  \State Model update: $\vtheta_{t+1}\leftarrow \nabla_{\theta} \gL(\vtheta_t, \gB)$ 
  \vspace{0.1em}
  \If{$\big(t\ge t_0$ \textbf{and} $t\bmod L=0\big)$}
    \State $r(x)\leftarrow$ Gaussianized rank of $y(x)$ \Comment{Eq.~\ref{eq:rankgauss}}
    \State $\mA\leftarrow\gamma\mA+\sum_{x\in\gS_t}\tphi(x)\tphi(x)^\top$,\quad $\vb\leftarrow\gamma\vb+\sum_{x\in\gS_t}r(x)\,\tphi(x)$ \Comment{Eq.~\ref{eq:ridge}}
  \EndIf
  \If{$t\ge t_0$ \textbf{and} $t\bmod U=0$}
    \State $\vw\leftarrow(\mA+\bar\lambda\mI)^{-1}\vb$ \Comment{\textit{Selector Update}}
  \EndIf
\EndFor
\end{algorithmic}
\end{algorithm}
\subsection{An online linear scorer on a frozen embedding}
\label{sec:scorer}
We hereby introduce the fitting and inference steps of the online linear data scorer, which is formalized as a linear vector $\vw_t$.
Given the frozen embedding $\vphi(x)\in\R^d$, the score of candidate $x$ is predicted by a simple linear transformation
\begin{equation}
s_t(x)=\vw_t^\top\tphi(x),\qquad \tphi(x)=\big[\vphi(x);\,1\big]\in\R^{d+1}.
\label{eq:score}
\end{equation}
The document embeddings are pre-computed once for the whole training corpus, which are frozen throughout training. The linear scorer $\vw_t$ is refitted to the \textbf{\textit{gaussianized ranks}} of the documents along their exact utility scores. Therefore, the scorer learns adaptive data preference from the language model through its evolving gradients.

\paragraph{Gaussianized ranks.}\label{sec:gaussian_ranks} Since the raw utility scores are computed from high-dimensional gradient vectors, they can contain outliers with extreme values, which can collapse the prediction of the linear scorer. We therefore fit the scorer $\vw_t$ to the Gaussianized rank of each document within the current training batch.   
Denote $\gS_s$ as the documents the scorer learns from on labeled step $s$, $n_s=|\gS_s|$, and $y_s(x)$ the label of $x\in\gS_s$. The Gaussianized rank is a transformation of document ranks on utility scores $\mathrm{rank}_s(x)\in\{1,\dots,n_s\}$ upon a Gaussian distribution,
\begin{equation}
r_s(x)=\Phi^{-1}\!\Big(\frac{\mathrm{rank}_s(x)-\tfrac12}{n_s}\Big),
\label{eq:rankgauss}
\end{equation}
where $\Phi^{-1}$ is the standard normal quantile function. 
We adopt Gaussianized ranks mainly for robustness. As shown in Figure~\ref{fig:targets}, the scale of raw utilities drifts with large variances and heavy tails, where the most extreme outlier carries $\approx 15\%$ of a step's variance under raw utilities. 
In comparison, applying Gaussianized ranks perfectly smooths the score distribution onto a normal distribution. 


\paragraph{Running ridge.} With decay $\gamma$ per labeled step, and the sums running over the labeled steps $s\le t$ from the scorer's start step $t_0$ on,
\begin{equation}
\mA_t=\sum_{s\le t}\gamma^{\,t-s}\sum_{x\in\gS_s}\tphi(x)\tphi(x)^\top,\qquad
\vb_t=\sum_{s\le t}\gamma^{\,t-s}\sum_{x\in\gS_s}r_s(x)\,\tphi(x),\qquad
\vw_t=\big(\mA_t+\bar\vp_t\mI\big)^{-1}\vb_t ,
\label{eq:ridge}
\end{equation}
where $\bar\vp_t$ is a fixed fraction of the mean eigenvalue of the embedding block of $\mA_t$ and the bias is not regularized. Rank 0 updates the sufficient statistics from the labels of all ranks, solves the system every $U$ steps and broadcasts $\vw_t$.

\subsection{Selection from a candidate window}
\label{sec:select}
At each step, candidates are drawn from the corpus in streaming until the candidate window holds $\lambda \times$ the batch's tokens. 
We name $\lambda$ as the \emph{window factor}, which controls the selectivity of \method. 
The linear selector then scores every candidate, and selects the batch of top-ranked candidate documents with the highest estimated utility scores to serve in LLM training. We score on each individual document instead of sequence chunks to avoid noise from truncations.
We start training the linear selector after a warmup stage $t_{0}$, when the gradient-based utility labels are getting stable. Algorithm~\ref{alg:companion} summarizes the procedure. We provide pseudo-code for implementation in Appendix \S\ref{app:impl}.
\begin{figure*}[t!]
\centering
\includegraphics[width=\textwidth]{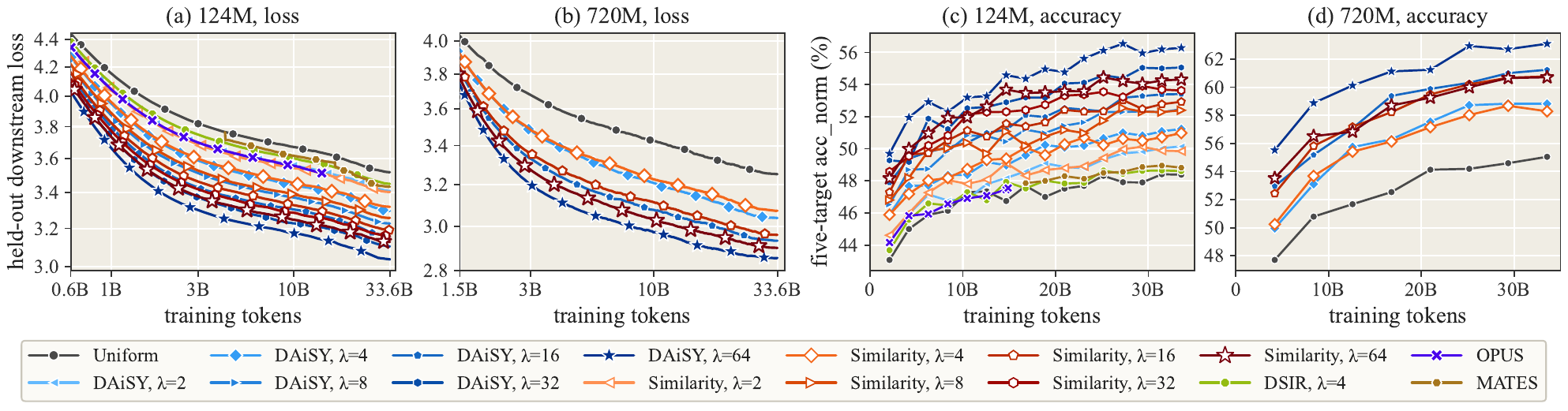}
\caption{\textbf{Downstream loss and zero-shot benchmark accuracy curves on target-oriented pretraining} at 124M and 720M. \method outperforms all prior data selection baselines. \textit{Similarity} works comparable as \method with small candidate window $\lambda=2, 4$, while its marginal benefits plateau with large $\lambda=16, 32, 64$.}
\label{fig:task_orient}
\vspace{-1em}
\end{figure*}
\vspace{-0.5em}
\section{Experiments}
\label{sec:results}
\vspace{-0.5em}
We train decoder-only transformers following Llama architecture~\citep{touvron2023llama} at $124$M and $720$M on FineWeb~\citep{penedo2024fineweb} with document-masked attention. Each run is trained on $33.6$B tokens, at around $15\times$ and $2.4\times$ of the model's Chinchilla-optimal \citet{hoffmann2022chinchilla} token budgets.
The embeddings $\vphi(x)$ are pre-computed for each document using XLM-RoBERTa-base~\citep{conneau2020xlmr} with 768-dimentions. We apply AdamW~\citep{loshchilov2019adamw} optimizer with a peak learning rate of $3\times10^{-4}$ with cosine scheduler.

\textbf{\method settings.} Unless stated otherwise, we adopt the default set of hyperparameters $L=4$, $R=200$, $U=10$, $\gamma=0.999$, and the candidate window factor $\lambda\in\{2,4,8,16,32,64\}$. The online selection starts at step $2500$.
We evaluate \method with various targets: On \textbf{\textit{task-oriented pretraining}} setup (\S\ref{sec:benchmark-target}), 
the target set is uniformly sampled from five \textit{reasoning benchmarks}: ARC-Easy and ARC-Challenge~\citep{clark2018arc}, PIQA~\citep{bisk2020piqa}, SciQ~\citep{welbl2017sciq} and HellaSwag~\citep{zellers2019hellaswag}. Only the \emph{training} split of each benchmark is used, so \textit{no test samples are leaked into training}. Each item consists of the question followed by its correct answer;
On \textbf{\textit{web-targeted}} setup (\S\ref{sec:webtargets}), we craft a high-quality subset from FineWeb-Edu \citep{penedo2024fineweb} as the target, which prevents \method from seeing any items from downstream benchmarks. Appendix~\ref{app:webtargets} provides more details.

\textbf{Baselines.} We compare \method to a series of baselines as follows:  
(1) \emph{\textbf{Uniform}} sampling trains the model with random streaming; 
(2) \emph{\textbf{Similarity}} scores a candidate by the cosine between $\vphi(x)$ and the mean embedding from the target $\vphi(\val)$; 
(3) \emph{\textbf{DSIR}}~\citep{xie2023dsir} scores a candidate by the log importance weight of DSIR's hashed n-gram model;
(4) \emph{\textbf{OPUS}}~\citep{wang2026opus} selects documents with their exact utilities from the LLM, while scoring only the first 512 tokens of each candidate document. We stopped OPUS at around $100$ GPU hours, with around $15$B token budget;
(5) \emph{\textbf{MATES}}~\citep{yu2024mates} fine-tunes a BERT influence model to predict target loss reductions as utility labels every 32k steps. The tuned BERT selector is then applied to score $\lambda=5\times$ tokens for the next $32k$ steps and select $20\%$ for training. We start the scoring at 2.5k step to keep consistent as our setup, and adopt the other hyperparameters following the default setup from the paper.

\textbf{Evaluation.} We report model performance in held-out losses and benchmark accuracies: \emph{\textbf{Downstream loss}} evaluates the average token losses on the held-out splits of the benchmarks (test split from ARC, PIQA, SciQ and validation split from HellaSwag); 
\textbf{\textit{Benchmark accuracy}} measures the zero-shot accuracy using lm-eval-harness~\citep{gao2023lmeval} on five targeted benchmarks and on three out-of-domain benchmarks that selection never targets,  WinoGrande~\citep{sakaguchi2020winogrande}, OpenBookQA~\citep{mihaylov2018openbookqa} and LAMBADA~\citep{paperno2016lambada}.
We also evaluate on the held-out \textbf{\textit{FineWeb loss}} and held-out \textbf{\textit{FineWeb-Edu loss}} in web-targeted experiments (\S\ref{sec:webtargets}). None of the data used for evaluation is included in the training corpus.

\begin{wrapfigure}{r}{0.56\textwidth}
\centering
\vspace{-0.5em}
\includegraphics[width=\linewidth]{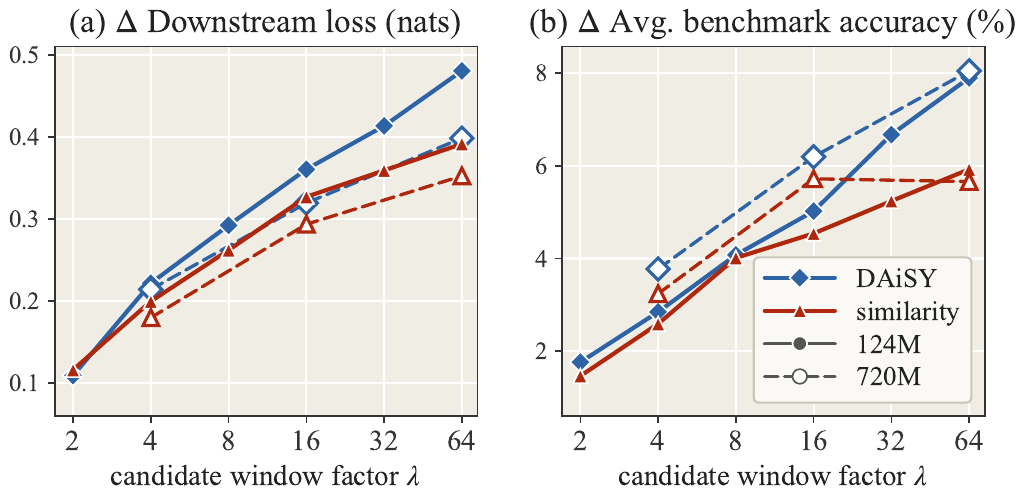}
\caption{\small \textbf{Gain over uniform sampling per candidate window}. \textbf{(a)}~Downstream loss improvements; \textbf{(b)}~Average zero-shot accuracy improvements against uniform baseline.}
\label{fig:lambdagain}
\vspace{-1.2em}
\end{wrapfigure}
\subsection{\method under Task-oriented Pretraining}
\label{sec:benchmark-target}
\begin{table}[t]
\centering
\caption{\textbf{Final downstream loss (\textit{Loss}) and zero-shot accuracies (\%) with task-oriented target}. \textbf{\textit{I.D. tasks}} reports the average accuracy across five benchmarks used to construct target set $\val$; \textbf{\textit{All tasks}} reports the average accuracy across all eight benchmarks including 3 out-of-domain tasks. Across various candidate window factor $\lambda$, \method keeps its leading position compared to Similarity and the Uniform baseline.}
\vspace{0.3em}
\adjustbox{max width=0.90\textwidth}{
\label{tab:task_orient}
\footnotesize
\setlength{\tabcolsep}{2.6pt}
\begin{tabular}{lllccccccccccc}
\toprule
& & & & \multicolumn{5}{c}{Benchmark-target tasks} & \multicolumn{3}{c}{Unseen tasks} & \multicolumn{2}{c}{Means} \\
\cmidrule(lr){5-9}\cmidrule(lr){10-12}\cmidrule(lr){13-14}
$\lambda$ & Scorer & Target & Loss & ARC-E & ARC-C & PIQA & SciQ & HellaSw. & WG & OBQA & LAMB. & I.D. tasks & All tasks \\
\midrule
\multicolumn{14}{l}{\emph{124M}} \\
\multicolumn{3}{l}{Uniform} & 3.521 & 47.01 & 21.59 & 67.30 & 79.70 & 32.73 & 53.04 & 17.00 & 33.77 & 48.38 & 43.22 \\
\midrule
5 & MATES & \textit{Bench.} & 3.435 & 49.24 & 19.37 & 66.97 & 79.10 & 32.93 & 52.09 & 17.40 & 35.32 & 48.82 & 43.61 \\
\midrule
\multirow{3}{*}{4} & DSIR & \textit{Bench.} & 3.452 & 49.28 & 21.42 & 68.50 & 77.80 & 32.38 & \textbf{54.38} & 18.40 & 28.93 & 48.61 & 43.09 \\
& Similarity & \textit{Bench.} & 3.321 & 52.90 & \textbf{22.27} & \textbf{70.78} & 80.50 & 34.43 & 52.17 & \textbf{21.00} & 32.87 & 50.97 & 45.11 \\
 & \cellcolor{daisyrow}\method & \cellcolor{daisyrow}\textit{Bench.} & \cellcolor{daisyrow}\textbf{3.299} & \cellcolor{daisyrow}\textbf{53.45} & \cellcolor{daisyrow}\textbf{22.27} & \cellcolor{daisyrow} 69.91 & \cellcolor{daisyrow}\textbf{81.00} & \cellcolor{daisyrow}\textbf{34.52} & \cellcolor{daisyrow}50.36 & \cellcolor{daisyrow}18.80 & \cellcolor{daisyrow}\textbf{35.90} & \cellcolor{daisyrow}\textbf{51.23} & \cellcolor{daisyrow}\textbf{45.15} \\
\midrule
\multirow{2}{*}{16} & Similarity & \textit{Bench.} & 3.194 & 57.87 & 24.83 & \textbf{71.98} & 81.70 & 34.01 & 49.72 & 20.80 & 29.75 & 52.92 & 45.61 \\
 & \cellcolor{daisyrow}\method & \cellcolor{daisyrow}\textit{Bench.} & \cellcolor{daisyrow}\textbf{3.160} & \cellcolor{daisyrow}\textbf{59.05} & \cellcolor{daisyrow}\textbf{25.17} & \cellcolor{daisyrow}71.33 & \cellcolor{daisyrow}82.40 & \cellcolor{daisyrow}\textbf{34.84} & \cellcolor{daisyrow}\textbf{52.01} & \cellcolor{daisyrow}\textbf{21.80} & \cellcolor{daisyrow}\textbf{34.62} & \cellcolor{daisyrow}\textbf{53.41} & \cellcolor{daisyrow}\textbf{46.93} \\
\midrule
\multirow{2}{*}{64} & Similarity & \textit{Bench.} & 3.129 & 61.78 & 27.05 & 71.76 & 85.10 & 33.08 & 51.30 & 22.40 & 25.77 & 54.31 & 46.38 \\
 & \cellcolor{daisyrow}\method & \cellcolor{daisyrow}\textit{Bench.} & \cellcolor{daisyrow}\textbf{3.040} & \cellcolor{daisyrow}\textbf{62.92} & \cellcolor{daisyrow}\textbf{27.99} & \cellcolor{daisyrow}\textbf{72.91} & \cellcolor{daisyrow}\textbf{85.90} & \cellcolor{daisyrow}\textbf{35.23} & \cellcolor{daisyrow}\textbf{52.41} & \cellcolor{daisyrow}\textbf{23.60} & \cellcolor{daisyrow}\textbf{30.72} & \cellcolor{daisyrow}\textbf{56.28} & \cellcolor{daisyrow}\textbf{48.52} \\
\midrule
\midrule
\multicolumn{14}{l}{\emph{720M}} \\
\multicolumn{3}{l}{Uniform} & 3.255 & 55.43 & 23.98 & 72.36 & 84.90 & 40.02 & 54.14 & 19.80 & 42.87 & 55.05 & 49.01 \\
\midrule
\multirow{2}{*}{4} & Similarity & \textit{Bench.} & 3.075 & 61.99 & \textbf{27.90} & \textbf{75.30} & 86.90 & 41.46 & 53.91 & 23.00 & 44.65 & 58.29 & 51.63 \\
 & \cellcolor{daisyrow}\method & \cellcolor{daisyrow}\textit{Bench.} & \cellcolor{daisyrow}\textbf{3.040} & \cellcolor{daisyrow}\textbf{62.92} & \cellcolor{daisyrow}26.45 & \cellcolor{daisyrow}74.65 & \cellcolor{daisyrow}\textbf{88.00} & \cellcolor{daisyrow}\textbf{42.68} & \cellcolor{daisyrow}\textbf{54.06} & \cellcolor{daisyrow}\textbf{24.40} & \cellcolor{daisyrow}\textbf{46.63} & \cellcolor{daisyrow}\textbf{58.82} & \cellcolor{daisyrow}\textbf{52.40} \\
\midrule
\multirow{2}{*}{16} & Similarity & \textit{Bench.} & 2.961 & 65.99 & 30.03 & \textbf{76.88} & \textbf{89.00} & 40.66 & 52.17 & \textbf{24.60} & 38.93 & 60.77 & 52.44 \\
 & \cellcolor{daisyrow}\method & \cellcolor{daisyrow}\textit{Bench.} & \cellcolor{daisyrow}\textbf{2.935} & \cellcolor{daisyrow}\textbf{68.10} & \cellcolor{daisyrow}\textbf{30.12} & \cellcolor{daisyrow}75.79 & \cellcolor{daisyrow}88.30 & \cellcolor{daisyrow}\textbf{43.19} & \cellcolor{daisyrow}\textbf{54.22} & \cellcolor{daisyrow}\textbf{24.60} & \cellcolor{daisyrow}\textbf{45.10} & \cellcolor{daisyrow}\textbf{61.24} & \cellcolor{daisyrow}\textbf{53.77} \\
\midrule
\multirow{2}{*}{64} & Similarity & \textit{Bench.} & 2.902 & 68.86 & \textbf{33.36} & 76.22 & 88.60 & 39.03 & 53.75 & 24.60 & 32.85 & 60.71 & 51.84 \\
 & \cellcolor{daisyrow}\method & \cellcolor{daisyrow}\textit{Bench.} & \cellcolor{daisyrow}\textbf{2.856} & \cellcolor{daisyrow}\textbf{71.21} & \cellcolor{daisyrow}33.28 & \cellcolor{daisyrow}\textbf{76.82} & \cellcolor{daisyrow}\textbf{89.00} & \cellcolor{daisyrow}\textbf{42.40} & \cellcolor{daisyrow}\textbf{54.14} & \cellcolor{daisyrow}\textbf{25.60} & \cellcolor{daisyrow}\textbf{37.61} & \cellcolor{daisyrow}\textbf{63.10} & \cellcolor{daisyrow}\textbf{54.11} \\
\bottomrule
\end{tabular}}
\end{table}

\textbf{\method significantly improves data efficiency with task-oriented target.} 
Figure \ref{fig:task_orient} and Table \ref{tab:task_orient} report the downstream losses and zero-shot benchmark accuracies. At both model scales, \method outperforms all the other baselines by a significant margin. Specifically, \method outperforms MATES and DSIR at 124M scale. 
\textit{Similarity} is a strong baseline with small candidate window ($\lambda=2, 4$) using only embedding cosine distance, where \method only obtains a 0.3--0.5\% accuracy gain on average. However, the advantage of \method against \textit{Similarity} grows with larger candidate window ($\lambda=64$) with $2$-$3$\% gain in average in-domain accuracies.
Notably, \method achieves better accuracies on three out-of-domain tasks than \textit{Similarity} across all candidate window factors, which indicates \method can identify more sophisticated semantic affinity by learning from gradient-based utilities, rather than the plain representation similarities.
Compared with the uniform baseline, \method with $\lambda=4$ reaches the final downstream loss with $7.6\times$ fewer tokens at 124M and $4.7\times$ fewer at 720M which remarkably improves data efficiency under task-oriented targets.

\textbf{Larger candidate window earns downstream improvement with costs on general loss.} The window factor $\lambda$ controls the size of the candidate pool for selection, reflecting how \textit{selective} the data selector (\method, \textit{Similarity}) applied. According to Figure~\ref{fig:lambdagain}, the downstream gain against uniform baseline on both losses and accuracy grow with the candidate window at both model scales. Notably, the advantage from \method above \textit{Similarity} baseline is also augmented with increasing candidate window.
However, the caveat is that the downstream improvements come with a price paid on general FineWeb loss. Table \ref{tab:generality} reports the validation loss on a held-out set on FineWeb. Larger candidate window leads to increasingly severe FineWeb loss degradation with both online data selection methods, while the \textit{Similarity} selection leads to larger degradations than \method at $\lambda=64$.

\textbf{Great efficiency gain against prior online selection methods.} The gain from \method carries over to compute, whereas previous model-aware baselines OPUS and MATES fall behind uniform sampling (Figure~\ref{fig:efficiency_gpuh} in Appendix~\ref{sec:cost}). 
Standard uniform training completes the whole run in 28.8 GPU-hours with a downstream loss of $3.519$ at 124M, and \method costs 32.6 - 43.4 GPU-hours with candidate window $2$ to $64$ with superior improvements above uniform baselines. At this budget, OPUS has trained for only about 39k steps with downstream loss $3.625$. MATES reaches $3.573$ within the 28.8 GPUh budget, which spends 17.9 GPU-hours on BERT scorer tuning at stage boundaries.
OPUS and MATES would take 209 GPU hours and 76.8 GPU hours to complete the full 33.6B tokens at the 124M model scale, failing to outperform uniform training at equal compute. 

\begin{figure*}[t!]
\centering
\includegraphics[width=\textwidth]{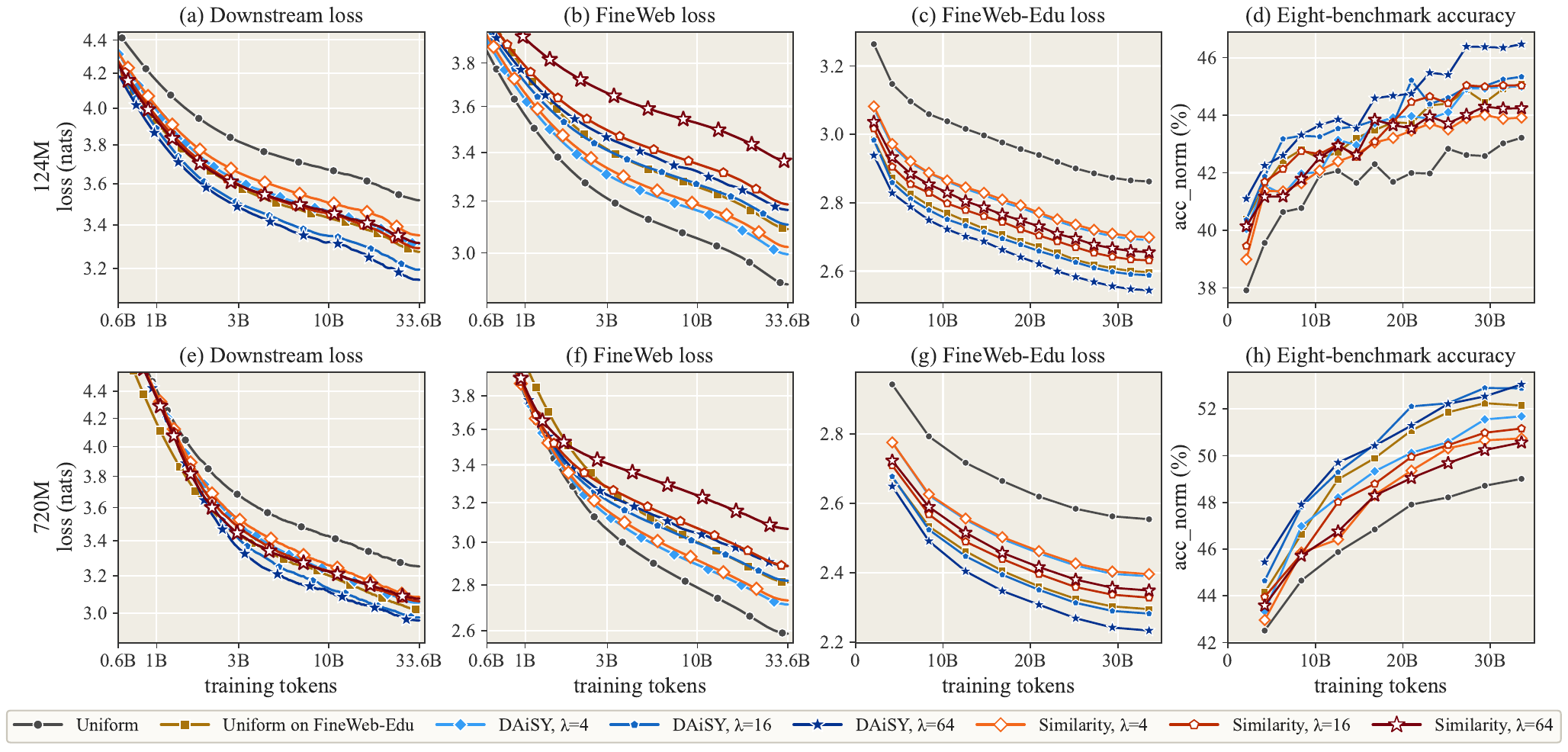}
\caption{\textbf{Held-out losses and zero-shot accuracies (\%) with HQ FineWeb-Edu target.} (a, b, e, f)~\textbf{Held-out downstream loss and held-out FineWeb loss} at each model scale. (c)~\textbf{FineWeb-Edu loss} on a held-out set unseen in the target. (d, h)~Zero-shot Accuracy averaged over the five benchmarks.}
\label{fig:edu_efficiency}
\vspace{-1em}
\end{figure*}
\vspace{-0.5em}
\subsection{Steering Online Data Selection with High-quality Web Data}
\label{sec:webtargets}
\vspace{-0.5em}
\textbf{Setup.} Selection guided by the task-oriented targets has an underlying risk of overfitting, with a great degradation on general web learning. 
We then test whether online selection maintains its power guided by target set $\val$ crawled from general web data.
We construct two target sets $\val$ from FineWeb and FineWeb-Edu \citep{penedo2024fineweb}: (1) The \emph{\textbf{FineWeb topic}} target consists of 976 held-out FineWeb documents uniformly sampled from the 16 topics. The topics are determined from K-Means clustering on the embeddings; (2) The \emph{\textbf{high-quality FineWeb-Edu}} target consists of 721 FineWeb-Edu documents~\citep{penedo2024fineweb} with a \textit{quality score} $\geq 4$. The \textit{quality scores} are assigned by FineWebEdu-classifier \footnote{\url{https://huggingface.co/HuggingFaceFW/fineweb-edu-classifier}}. 
As a strict reference, we add an oracle FineWeb-Edu baseline training on a curated set of FineWeb-Edu documents with \textit{quality score} $\geq 3$ ($124$B tokens in total). 
In addition to the two held-out losses on FineWeb and downstream benchmarks, we report the loss on a held-out set from high-quality FineWeb-Edu (\textit{quality score}$\geq4$) that is unseen in the FineWeb-Edu target set. 

\textbf{High-quality FineWeb-Edu target transfers to the benchmarks.} 
\method with the high-quality FineWeb-Edu target improves the benchmarks although it contains no benchmark item. According to Table~\ref{tab:webtargets} and Figure~\ref{fig:edu_efficiency}, both \method and \textit{Similarity} significantly outperform Uniform baseline on FineWeb. 
Notably, \textbf{\textit{with the high-quality FineWeb-Edu target, the gap between the static and the learned scorer widens. }}Compared to the FineWeb-Edu baseline, all \textit{Similarity} runs fall behind on downstream loss and benchmark accuracies while \method with $\lambda=16$ and $64$ significantly outperform the strict FineWeb-Edu baseline with $0.7$-$1.5$\% average benchmark accuracy improvements at both 124M and 720M. 
Additionally, \method with high-quality FineWeb-Edu target shows marginal degradation on FineWeb loss compared with FineWeb-Edu baseline. In comparison, selection by \textit{Similarity} will lead to a severe cost on the general FineWeb data, which hurts the model's generalizability.

\begin{figure*}[t!]
\centering
\vspace{-0.5em}
\includegraphics[width=0.85\linewidth]{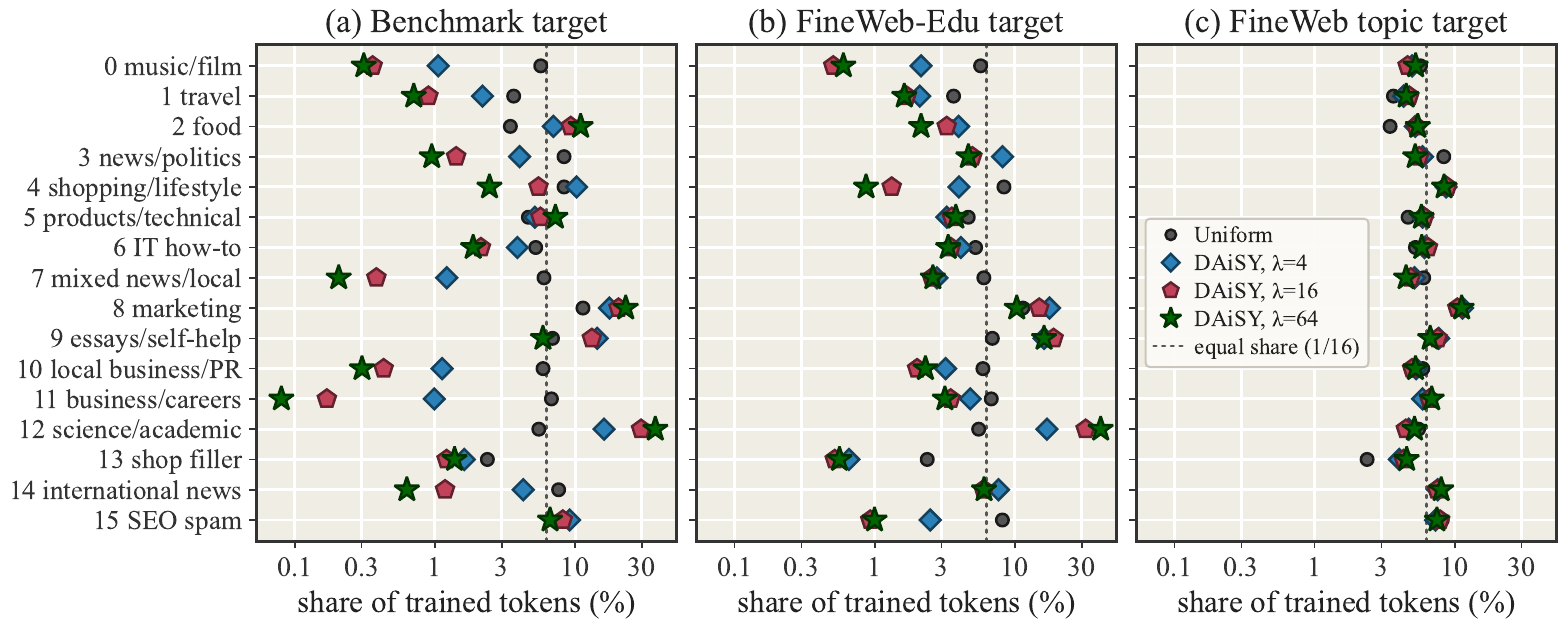}
\caption{\textbf{Topic composition of the trained documents at 124M}.}
\label{fig:selection_topics}
\vspace{-1.5em}
\end{figure*}
\textbf{FineWeb topic-uniform target gives no performance gain.} The FineWeb topic target does not improve upon uniform sampling on either FineWeb learning or downstream performance. \method with FineWeb topic target runs end with $0.013$-$0.033$ nats above Uniform baseline on FineWeb loss, $0.03$-$0.035$ nats degradation on FineWeb-Edu loss and achieve comparable benchmark accuracies across all candidate window sizes. 
Its selected document composition is close to a uniform sampling across topics. 
We regard these runs as a negative control: when the target resembles what uniform sampling already provides, selection by one-step utility cannot improve much upon uniform sampling.

\textbf{What data the targets select?}
Figure~\ref{fig:selection_topics} tracks which kind of documents are selected under different targets. 
With task-oriented benchmark target, selection upweights \textit{Science/academic}, \textit{Food}, and \textit{Marketing topics}, while largely concentrates on Science/academic, with proportion increasing with candidate window $16\rightarrow30\rightarrow37\%$ with $\lambda=4\rightarrow16\rightarrow64$.
Similarly, the high-quality FineWeb-Edu target also upweights the \textit{Science/academic} with $17$-$41$\%, while it also puts more weights on the \textit{essays/self-help} topic and effectively eliminates the \textit{SEO-spam} and \textit{shop filler} topics, which are considered as harmful and noisy sources.
In comparison, the FineWeb topic target keeps every topic around its natural share in uniform sampling, which explains its lack of any gain. The composition at 720M exhibits a similar selection preference, with details provided in Appendix~\ref{app:webtargets}.


\begin{table}[t!]
\centering
\caption{\textbf{Final losses and accuracies (\%) with web target}. \textbf{\textit{Down. Loss}} reports the average held-out loss across five benchmarks used in Table \ref{tab:task_orient}; \textbf{\textit{FW}}/\textbf{\textit{FW-Edu Loss}} reports the held-out loss on FineWeb and FineWeb-Edu. \textbf{\textit{Average}} reports the average zero-shot accuracy across all eight benchmarks. With FineWeb-Edu as target, \textit{Similarity} fails to outperform training on FineWeb-Edu target set, while \method can further improve the held-out loss on FineWeb-Edu and benchmark accuracies with $\lambda=16$ and $64$.}
\vspace{0.3em}
\adjustbox{max width=0.95\textwidth}{
\label{tab:webtargets}
\footnotesize
\setlength{\tabcolsep}{2.6pt}
\begin{tabular}{lllcccccccccccc}
\toprule
& & & \multicolumn{3}{c}{Loss}& \multicolumn{8}{c}{Downstream benchmarks} &  \\
\cmidrule(lr){4-6}\cmidrule(lr){7-14}
$\lambda$ & Scorer & Target & FW & FW-Edu & Down. & ARC-E & ARC-C & PIQA & SciQ & HellaSw. & WG & OBQA & LAMB. & Average\\
\midrule
\multicolumn{13}{l}{\emph{124M}} \\
\multicolumn{3}{l}{Uniform} & 2.887 & 2.862 & 3.521 & 47.01 & 21.59 & 67.30 & 79.70 & 32.73 & 53.04 & 17.00 & 33.77 & 43.22 \\
\multicolumn{3}{l}{FineWeb-Edu} & 3.093 &  2.597 &  3.277 & 57.62 & 24.15 & 65.94 & 82.80 & 33.10 & 52.49 & 21.20 & 29.40 & 45.09 \\
\midrule
\multirow{3}{*}{4} & Similarity & HQ FineWeb-Edu & 3.024 & 2.699 & 3.352 & 51.60 & 22.01 & 67.46 & 79.90 & 33.11 & 51.54 & \textbf{20.80} & 32.29 & 43.92 \\
 & \cellcolor{daisyrow}\method & \cellcolor{daisyrow}FineWeb topics & \cellcolor{daisyrow} 2.900 & \cellcolor{daisyrow} 2.900 & \cellcolor{daisyrow}3.550 & \cellcolor{daisyrow}45.20 & \cellcolor{daisyrow}20.14 & \cellcolor{daisyrow}\textbf{69.10} & \cellcolor{daisyrow}77.60 & \cellcolor{daisyrow}33.36 & \cellcolor{daisyrow}\textbf{53.20} & \cellcolor{daisyrow}17.80 & \cellcolor{daisyrow}\textbf{34.43} & \cellcolor{daisyrow}43.55 \\
 & \cellcolor{darkdaisyrow}\method & \cellcolor{darkdaisyrow}HQ FineWeb-Edu & \cellcolor{darkdaisyrow}2.997 &  \cellcolor{darkdaisyrow}\textbf{2.691} &\cellcolor{darkdaisyrow}\textbf{3.311} & \cellcolor{darkdaisyrow}\textbf{53.41} & \cellcolor{darkdaisyrow}\textbf{23.12} & \cellcolor{darkdaisyrow}68.39 & \cellcolor{darkdaisyrow}\textbf{80.60} & \cellcolor{darkdaisyrow}\textbf{34.05} & \cellcolor{darkdaisyrow}51.22 & \cellcolor{darkdaisyrow}20.40 & \cellcolor{darkdaisyrow}34.08 & \cellcolor{darkdaisyrow}\textbf{45.00} \\
\midrule
\multirow{3}{*}{16} & Similarity & HQ FineWeb-Edu & 3.189 &  2.631 & 3.292 & 57.87 & 25.26 & 65.07 & \textbf{84.50} & 31.95 & \textbf{53.59} & \textbf{21.40} & 27.44 & 45.03 \\
 & \cellcolor{daisyrow}\method & \cellcolor{daisyrow}FineWeb topics & \cellcolor{daisyrow}2.906 & \cellcolor{daisyrow}2.916 & \cellcolor{daisyrow}3.572 & \cellcolor{daisyrow}45.66 & \cellcolor{daisyrow}21.25 & \cellcolor{daisyrow}\textbf{68.44} & \cellcolor{daisyrow}76.70 & \cellcolor{daisyrow}33.43 & \cellcolor{daisyrow}52.41 & \cellcolor{daisyrow}16.00 & \cellcolor{daisyrow}\textbf{33.61} & \cellcolor{daisyrow}42.73 \\
 & \cellcolor{darkdaisyrow}\method & \cellcolor{darkdaisyrow}HQ FineWeb-Edu & \cellcolor{darkdaisyrow}3.110  & \cellcolor{darkdaisyrow}\textbf{2.588} & \cellcolor{darkdaisyrow}\textbf{3.195} & \cellcolor{darkdaisyrow}\textbf{59.26} & \cellcolor{darkdaisyrow}\textbf{25.43} & \cellcolor{darkdaisyrow}68.34 & \cellcolor{darkdaisyrow}83.40 & \cellcolor{darkdaisyrow}\textbf{33.61} & \cellcolor{darkdaisyrow}49.72 & \cellcolor{darkdaisyrow}21.00 & \cellcolor{darkdaisyrow}28.45 & \cellcolor{darkdaisyrow}\textbf{45.34} \\
\midrule
\multirow{3}{*}{64} & Similarity & HQ FineWeb-Edu &  3.364 & 2.655 & 3.317 & \textbf{62.96} & \textbf{28.24} & 63.82 & \textbf{85.90} & 30.22 & 51.54 & 22.00 & 22.32 & 44.24 \\
 & \cellcolor{daisyrow}\method & \cellcolor{daisyrow}FineWeb topics & \cellcolor{daisyrow}2.920 & \cellcolor{daisyrow}2.904 & \cellcolor{daisyrow}3.568 & \cellcolor{daisyrow}44.61 & \cellcolor{daisyrow}21.16 & \cellcolor{daisyrow}\textbf{68.39} & \cellcolor{daisyrow}78.50 & \cellcolor{daisyrow}\textbf{33.57} & \cellcolor{daisyrow}53.35 & \cellcolor{daisyrow}16.60 & \cellcolor{daisyrow}\textbf{33.11} & \cellcolor{daisyrow}43.08 \\
 & \cellcolor{darkdaisyrow}\method & \cellcolor{darkdaisyrow}HQ FineWeb-Edu & \cellcolor{darkdaisyrow}3.166 & \cellcolor{darkdaisyrow}\textbf{2.544} & \cellcolor{darkdaisyrow}\textbf{3.150} & \cellcolor{darkdaisyrow}62.42 & \cellcolor{darkdaisyrow}26.02 & \cellcolor{darkdaisyrow}68.23 & \cellcolor{darkdaisyrow}84.70 & \cellcolor{darkdaisyrow}33.02 & \cellcolor{darkdaisyrow}\textbf{53.67} & \cellcolor{darkdaisyrow}\textbf{22.80} & \cellcolor{darkdaisyrow}28.16 & \cellcolor{darkdaisyrow}\textbf{46.47} \\
\midrule
\multicolumn{13}{l}{\emph{720M}} \\
\multicolumn{3}{l}{Uniform} & 2.590 &  2.554 & 3.255 & 55.43 & 23.98 & 72.36 & 84.90 & 40.02 & 54.14 & 19.80 & 42.87 & 49.01 \\
\multicolumn{3}{l}{FineWeb-Edu} & 2.814 & 2.295 & 3.022 & 67.97 & 31.48 & 70.84 & 89.40 & 39.41 & 55.09 & 25.80 & 37.94 & 52.15 \\
\midrule
\multirow{2}{*}{4} & Similarity & HQ FineWeb-Edu & 2.733  & 2.396 & 3.086 & 62.84 & \textbf{27.82} & 70.84 & 86.00 & 40.20 & 52.96 & \textbf{23.80} & 42.17 & 50.74 \\
 & \cellcolor{darkdaisyrow}\method & \cellcolor{darkdaisyrow}HQ FineWeb-Edu & \cellcolor{darkdaisyrow}2.715  & \cellcolor{darkdaisyrow}\textbf{2.390} & \cellcolor{darkdaisyrow}\textbf{3.055} & \cellcolor{darkdaisyrow}\textbf{63.68} & \cellcolor{darkdaisyrow}27.13 & \cellcolor{darkdaisyrow}\textbf{72.91} & \cellcolor{darkdaisyrow}\textbf{87.20} & \cellcolor{darkdaisyrow}\textbf{41.81} & \cellcolor{darkdaisyrow}\textbf{55.33} & \cellcolor{darkdaisyrow}22.60 & \cellcolor{darkdaisyrow}\textbf{42.42} & \cellcolor{darkdaisyrow}\textbf{51.68} \\
\midrule
\multirow{2}{*}{16} & Similarity & HQ FineWeb-Edu & 2.889 & 2.328 & 3.065 & \textbf{67.85} & 31.74 & 69.70 & \textbf{88.50} & 37.62 & 53.35 & 25.80 & 36.77 & 51.16 \\
 & \cellcolor{darkdaisyrow}\method & \cellcolor{darkdaisyrow}HQ FineWeb-Edu & \cellcolor{darkdaisyrow}2.821 & \cellcolor{darkdaisyrow}\textbf{2.283} & \cellcolor{darkdaisyrow}\textbf{2.979} & \cellcolor{darkdaisyrow}67.21 & \cellcolor{darkdaisyrow}\textbf{33.62} & \cellcolor{darkdaisyrow}\textbf{72.63} & \cellcolor{darkdaisyrow}\textbf{88.50} & \cellcolor{darkdaisyrow}\textbf{40.18} & \cellcolor{darkdaisyrow}\textbf{53.75} & \cellcolor{darkdaisyrow}\textbf{27.80} & \cellcolor{darkdaisyrow}\textbf{38.72} & \cellcolor{darkdaisyrow}\textbf{52.88} \\
\midrule
\multirow{2}{*}{64} & Similarity & HQ FineWeb-Edu & 3.068 & 2.349 & 3.077 & 70.33 & 34.90 & 67.30 & 88.30 & 35.08 & \textbf{53.20} & 25.60 & 32.02 & 50.57 \\
 & \cellcolor{darkdaisyrow}\method & \cellcolor{darkdaisyrow}HQ FineWeb-Edu & \cellcolor{darkdaisyrow}2.888 & \cellcolor{darkdaisyrow}\textbf{2.234} & \cellcolor{darkdaisyrow}\textbf{2.964} & \cellcolor{darkdaisyrow}\textbf{71.09} & \cellcolor{darkdaisyrow}\textbf{35.32} & \cellcolor{darkdaisyrow}\textbf{71.06} & \cellcolor{darkdaisyrow}\textbf{90.50} & \cellcolor{darkdaisyrow}\textbf{38.84} & \cellcolor{darkdaisyrow}52.96 & \cellcolor{darkdaisyrow}\textbf{29.20} & \cellcolor{darkdaisyrow}\textbf{36.89} & \cellcolor{darkdaisyrow}\textbf{53.05} \\
\bottomrule
\end{tabular}}
\vspace{-1em}
\end{table}
\vspace{-0.5em}
\subsection{Could the Pretraining Gains Survive Instruction-tuning?}
\label{sec:sft}
\vspace{-0.5em}

\begin{wrapfigure}{r}{0.54\textwidth}
\centering
\vspace{-1.5em}
\centering
\includegraphics[width=\linewidth]{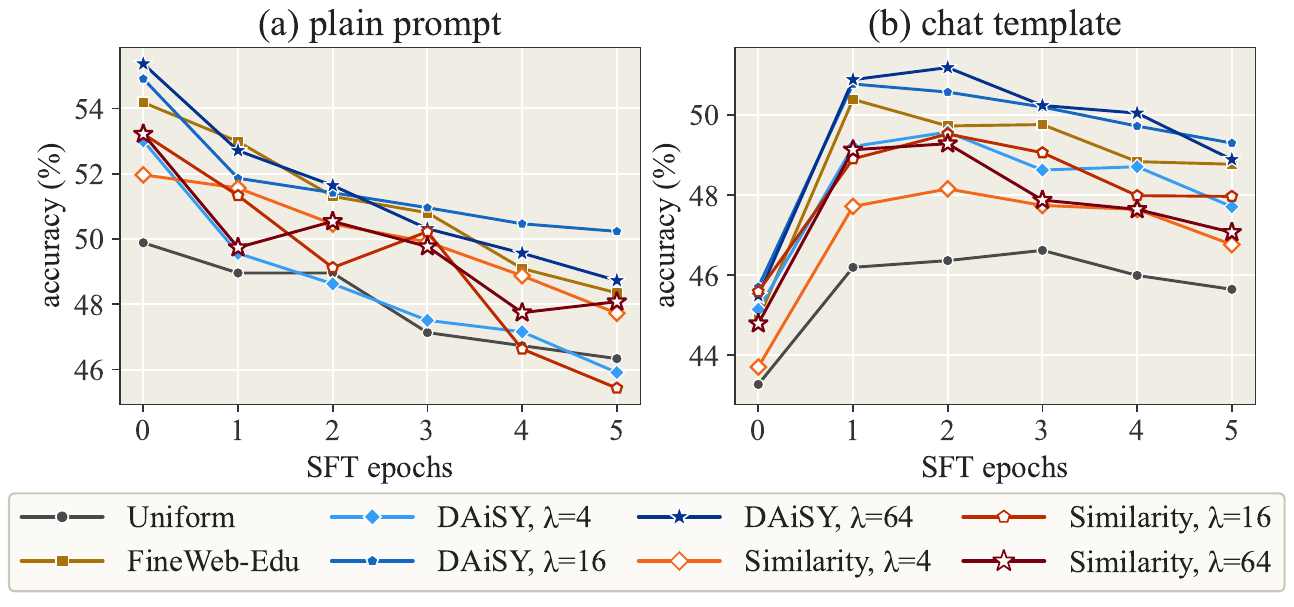}
\vspace{-1.5em}
\caption{\small \textbf{Zero-shot accuracy after $k$ epochs of instruction tuning at 720M}, averaged over the seven reasoning benchmarks. (a) \textbf{\textit{plain prompts}}; (b) \textbf{\textit{Tulu chat template}}. Epoch 0 denotes pretrained models.}
\label{fig:sft_means}
\vspace{-1.5em}
\end{wrapfigure}
Recent literature reported that a pretraining gain on benchmark-like text may limit the benefits from post-training \citep{springer2025overtrainedlanguagemodelsharder}. To evaluate \method's cross-stage advantage, we fine-tune 720M checkpoints on Tulu~3 SFT mixture~\citep{lambert2024tulu3} for one to five epochs with AdamW and the WSD \citep{hägele2024scalinglawscomputeoptimaltraining} scheduler at learning rate $10^{-4}$. At the end of each epoch, we anneal a copy of the model to zero learning rate, which ensures that the model after $k$ epochs made a complete $k$-epoch tuning. We evaluate every model zero-shot in two formats: with \textit{plain prompts} as in pretraining and with Tulu's \textit{chat template}, where each question is wrapped as user-assistant turns and the answer is scored as the assistant's reply. More details are provided in Appendix~\ref{app:sft}.

\textbf{The lead of \method survives tuning.} Figure~\ref{fig:sft_means} and Table~\ref{tab:sft_epochs} report the reasoning accuracies evolving during multi-epoch instruction tuning. With the chat format, which matches the format of instruction-tuning data, every \method run leads uniform baseline after every epoch by $2.0$--$4.8\%$ in average accuracy.
Notably, compared with the strong FineWeb-Edu baseline, \method with $\lambda=16$ and $64$ consistently maintain their leads, with $0.39$ and $0.80$\% improvements on the best accuracy achieved across instruction tuning.
With \textit{plain template}, however, the accuracy decreases drastically on every run, while \method with $\lambda=16, 64$ still shows the least degradation compared to other baselines.

\subsection{Multilingual Pretraining with Online Data Selection}
\label{sec:multilingual}
\paragraph{Setup.} We further test \method's capability over a multilingual pool with more language-wise heterogeneity. 
We train 124M language models on six languages from FineWeb2~\citep{penedo2025fineweb2}, including French, German, Chinese, Spanish, Portuguese and Italian, which contains 2.37B documents and 2.27T tokens in total. 
We apply the multilingual BPE tokenizer with 200,064 vocabularies following Apertus \citep{apertus2025apertus} and mmBERT-base~\citep{marone2025mmbert} as multilingual embeddings $\vphi(x)$.
Similar to English, we test three targets to steer online selection: (1) \textbf{\textit{Multilingual Benchmark Target}}: consists of 6 languages' subsets from Global-MMLU~\citep{singh2024globalmmlu} and INCLUDE~\citep{romanou2025include}; the reported \textit{downstream loss} is the mean loss on the held-out items of these tasks; (2) \textbf{\textit{High-quality FineWeb2 Target}}: We curate the documents with high \textit{multilingual quality score} ($\geq3$) assigned by a multilingual-FineWeb classifier \citep{sabolčec2026qualityclassifiers}; (3) \textbf{\textit{Mixed target}}: combine the curated high-quality FineWeb2 and benchmark splits together.
Since the natural distribution from FineWeb2 is unbalanced language-wise, we run Uniform baseline, \textit{Similarity} and \method with two types of data streams: (1) \textbf{\textit{Natural language mix}}: follow the natural language composition in FineWeb2, where $50.1\%$ tokens are from Chinese while only $5.4\%$ from Italian; (2) \textbf{\textit{Balanced stream}}: we balance the composition from each language, i.e. $16.7\%$ from each of six languages.  
Appendix~\ref{app:multilingual} provides more details on the language composition and experimental configurations.

\begin{figure*}[t!]
\centering
\includegraphics[width=0.95\textwidth]{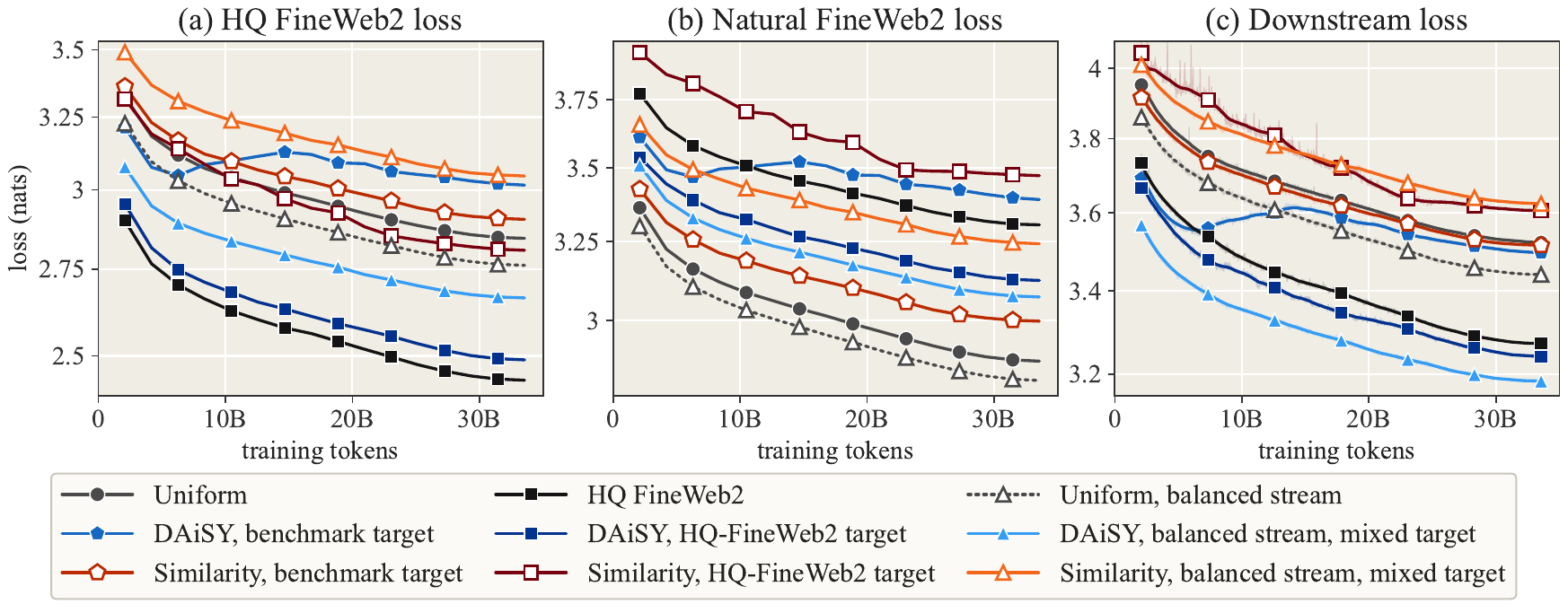}
\caption{\textbf{Multilingual pretraining with online data selection at $\lambda=16$.} (a)~\textbf{Held-out HQ FineWeb2 Loss}; (b)~\textbf{Held-out FineWeb2 Loss} following natural distribution. (c)~\textbf{Downstream loss}, average held-out loss on INCLUDE and Global-MMLU across 6 languages.}
\label{fig:multilingual}
\vspace{-2.5em}
\end{figure*}

\textbf{\method obtains downstream performance via benchmark target with an \textcolor{DarkRed}{observable overfitting}.} According to Figure \ref{fig:multilingual}, with multilingual benchmark target, \method with $\lambda=16$ effectively improves downstream losses in the first $5$B tokens while followed by a severe overfitting and ends up with a comparable performance as uniform baseline.
Figure \ref{fig:multilingual_lang} explains the overfitting: Since the benchmark-oriented selection is biased too heavily toward Chinese, all held-out losses on the other five languages degrade drastically while only Chinese realizes a continuous improvement.
Figure \ref{fig:fw2_lambda} provides multilingual benchmark-targeted \method with various candidate window $\lambda$, where $\lambda=4$ has not seen severe overfitting while the improvement is also limited; $\lambda=64$ performs a light overfitting around $10$B tokens, while getting recovered after $12$B tokens and ends up with the best improvements on averaged downstream performance. According to Table \ref{tab:multilingual}, compared with Uniform baseline with natural data stream, \method with $\lambda=4$ and $64$ improves both Global-MMLU ($0.04\%$ and $0.2\%$) and INCLUDE ($0.8\%$ and $2.5\%$); while $\lambda=16$ reaches comparable accuracies on both tasks. It indicates that \textit{choosing the right candidate window $\lambda$ is important with a narrow target from end tasks.}
Notably, \textit{Similarity} achieves NO benefit under multilingual selection. 

\textbf{The curated HQ-FineWeb2 target transfers to the multilingual benchmarks effectively.} 
With the HQ-FineWeb2 target at $\lambda=16$, \method consistently lowers the downstream loss and held-out HQ- and natural- FineWeb2 losses. According to Table \ref{tab:multilingual}, compared with the Uniform baseline with natural stream, \method achieves $0.28$ nats reduction on downstream losses, with $0.63\%$ and $2.5\%$ accuracy improvements on Global-MMLU and INCLUDE.
Compared with the strong baseline which directly trains on HQ-FineWeb2 (\textit{Uniform on HQ-FineWeb2}), $\lambda=16$ can still obtain $0.22\%$ improvement on INCLUDE, while Similarity performs \textcolor{DarkRed}{$-1.1\%$} and \textcolor{DarkRed}{$-1.6\%$} degradations on two benchmarks. 

\begin{wrapfigure}{r}{0.56\textwidth}
\centering
\vspace{-1.5em}
\includegraphics[width=\linewidth]{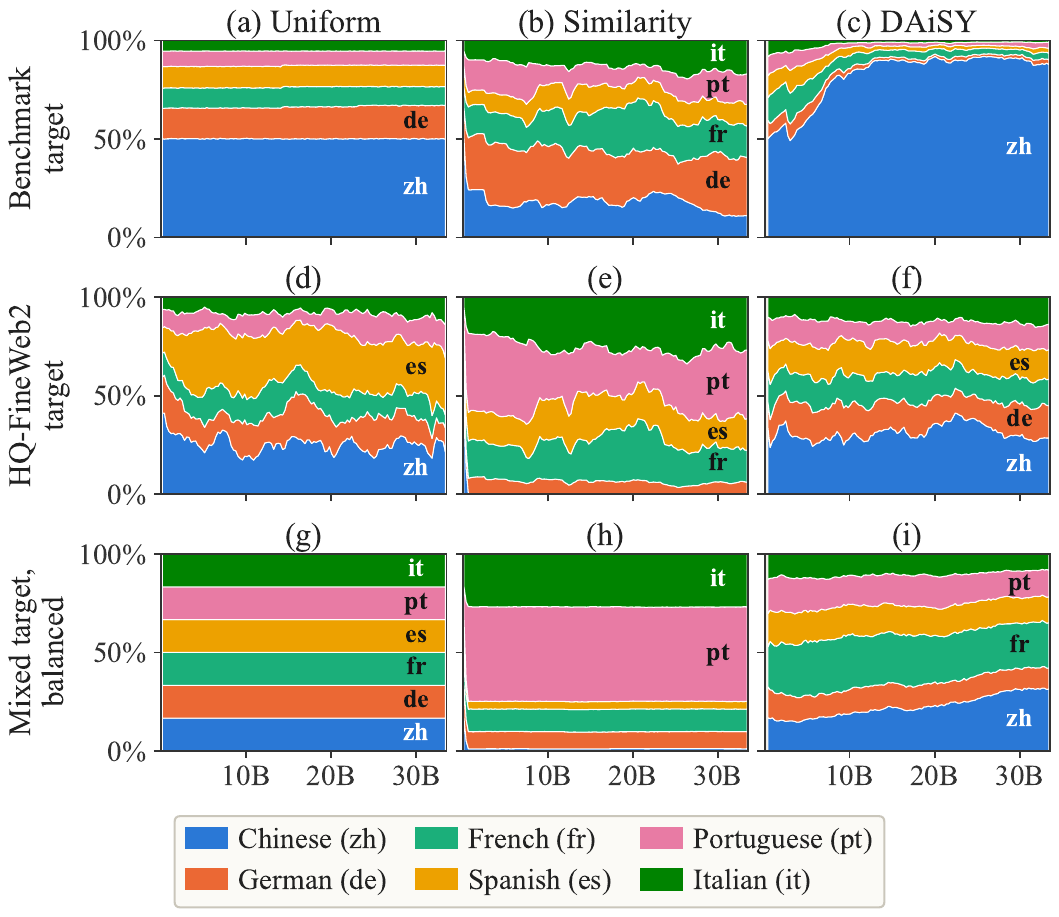}
\vspace{-1.5em}
\caption{\small \textbf{Language of the trained tokens at $\lambda=16$}. (a-c) Benchmark target with natural stream; (d-f) HQ-Fineweb2 target with natural stream; (g-i) Mixed target with balanced stream.}
\label{fig:fw2_langmix}
\vspace{-1.5em}
\end{wrapfigure}
\textbf{Balanced languages stream keeps the gain.} The natural FineWeb2 stream is strongly biased toward Chinese, we further experiment with balanced stream with uniform token shares across six langauges.
We apply a mix target, which combines the benchmark target and the curated HQ Fineweb2 set. 
The uniform baseline hereby is trained on the balanced stream, which obtains $0.05$--$0.08$ nats improvements on three held-out losses compared to uniform with natural language mix (Table~\ref{tab:multilingual}). 
Against the Uniform baseline with balanced stream, \method with $\lambda=16$ obtains a $0.68\%$ improvement on Global-MMLU and $0.29\%$ on INCLUDE, while \textit{Similarity} degrades both benchmark by \textcolor{DarkRed}{$-0.15\%$} and \textcolor{DarkRed}{$-1.75\%$}.
It is worth noting that \textit{\method with the mixed target and balanced language stream achieves the best downstream loss across all the other online selection runs} on multilingual pretraining.

\textbf{Which languages are selected more?} Figure~\ref{fig:fw2_langmix} presents the language composition under various selection runs at a candidate window $\lambda=16$. 
With natural stream, the uniform training is strongly biased toward Chinese. Using benchmark target, even though the target distribution is balanced across languages, \method augments the bias and selects $\approx80\%$ of tokens from Chinese;
The HQ-Fineweb2 target augments slightly on Spanish (es, $27.7\%$) and Chinese (zh, $25.2\%$). Guided by HQ-Fineweb2 target, \method keeps a more balanced distribution across languages, selecting $25$-$30\%$ from Chinese and around $20\%$ from Spanish and German.
On the balanced stream and mixed target, \method prioritizes Chinese and French most, with the share of Chinese smoothly increasing from $15.4\%$ to $31.5\%$.

In contrast, \textit{Similarity} selection moves away from Chinese in every setting. $17.6\%$ of its selected tokens are from Chinese with the benchmark target, while only $0.1\%$ are selected with the HQ-FineWeb2 target and $1.1\%$ with the mixed target on the balanced language stream. Instead, \textit{Similarity} prefers Portuguese with $48.0\%$ share under balanced language stream (Figure \ref{fig:fw2_langmix}). This leads to a strongly biased performance with only Portuguese improves smoothly while all the other languages perform severe degradations (Figure~\ref{fig:multilingual_lang}). 
It demonstrates that \textit{Similarity} can fail easily with a heterogeneous space and a less strong embedding model while \method stays effective.
\begin{figure*}[t]
\centering
\includegraphics[width=0.95\textwidth]{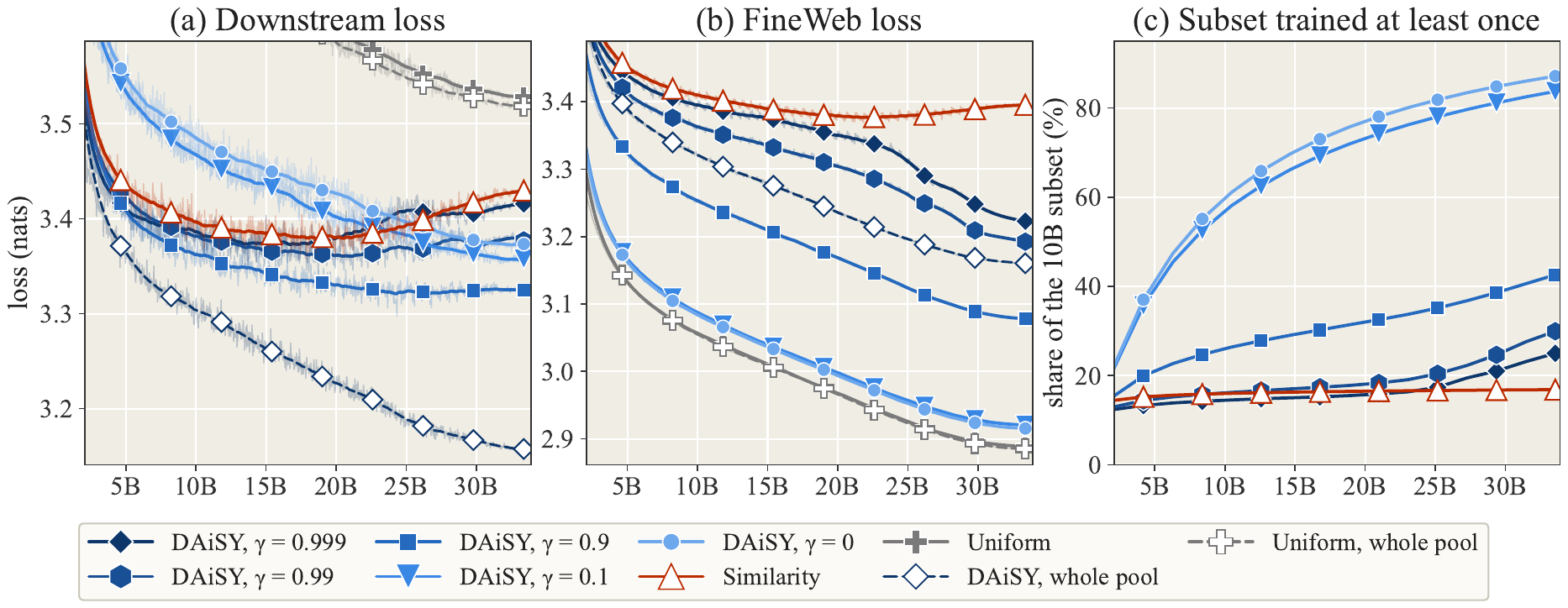}
\caption{\textbf{Selection from a 10B-token subset with $\lambda=16$.} \textbf{(a, b)}~Held-out downstream and FineWeb loss; \textbf{(c)}~Share of the subset trained at least once.}
\label{fig:budget_main}
\vspace{-1.5em}
\end{figure*}

\begin{wrapfigure}{r}{0.4\textwidth}
\centering
\vspace{-1.5em}
\includegraphics[width=\linewidth]{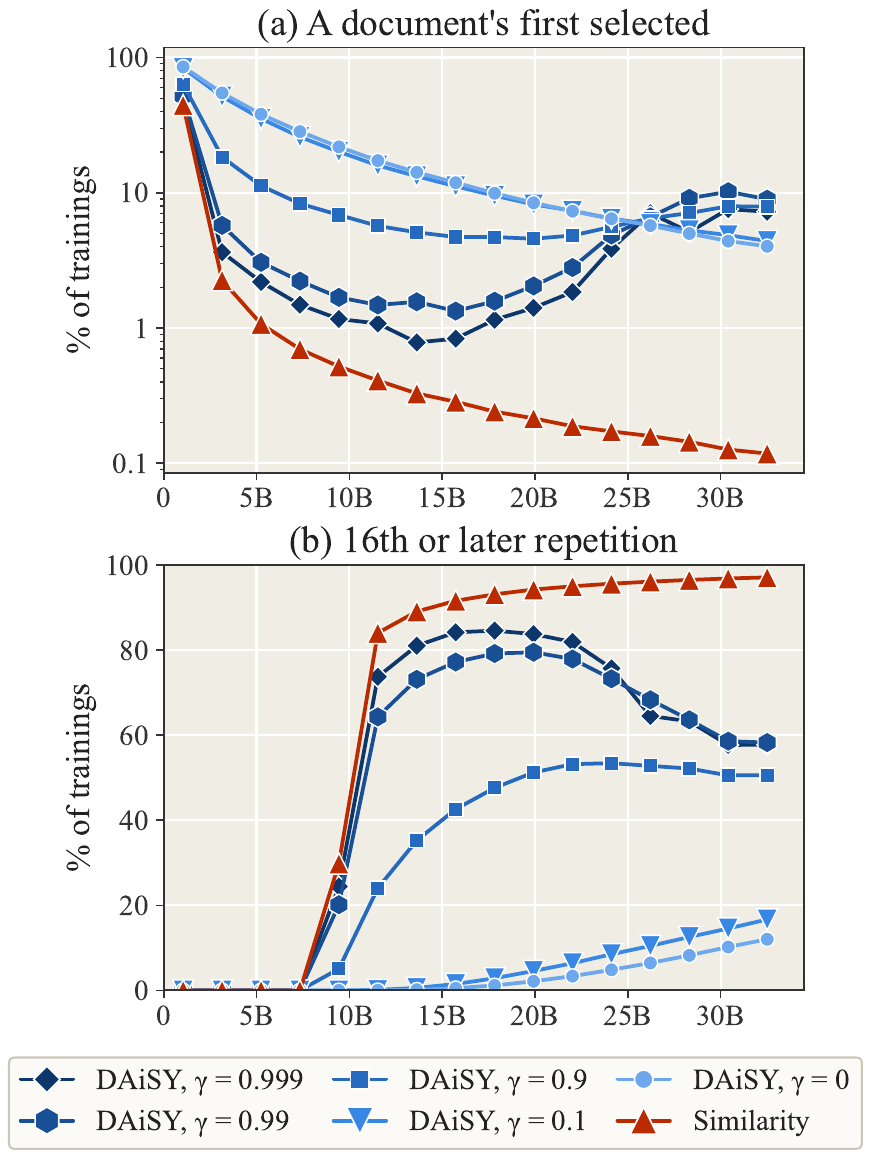}
\caption{\small \textbf{How the runs of Figure~\ref{fig:budget_main} spread their trainings}: \textbf{(a)}~a document's first seen or \textbf{(b)}~its 16th repetition or later.
}
\label{fig:repeat_lines}
\vspace{-2.5em}
\end{wrapfigure}
\vspace{-1em}
\section{What Happens on the Data Boundary?}
\label{sec:budget}
\vspace{-0.5em}
One expected advantage of an online selector is to identify the evolving data preference, especially when high-quality data runs out and the model requires more diversity. 
We simulate the data-constrained scenario by training the models on a $10$B subset from FineWeb, where the standard uniform training would therefore repeat each document by $3.36\times$, as the estimated optimal repetition times in \citep{muennighoff2025scalingdataconstrainedlanguagemodels}. We then train \method and the \textit{Similarity} baseline as the strongest static scorer with the same benchmark target as in \S\ref{sec:benchmark-target}.
\vspace{-1em}
\subsection{Long and Short \textit{Memory} of your Data Selector}
\vspace{-0.5em}
As introduced in \eqref{eq:ridge}, $\gamma$ controls the \textit{memory} of the data selector --- how much it depends on the model's historical data preference. 
With higher value (e.g. $\gamma=0.999$), the selector retains a strong dependence on previous steps, i.e. keeps a \textit{long memory}; with smaller value (e.g. $\gamma=0.1$), the selector memorizes less on history while adapting fast to the current data preference. 
In Figure \ref{fig:budget_main}, we compare \method with various \textit{memory factor} $\gamma$.
Comparing to uniform training on the whole FineWeb pool, the $10$B token limit with $3.36\times$ does not affect the performance, aligning with the results from \citep{muennighoff2025scalingdataconstrainedlanguagemodels}. 
However, the benefits from online selection significantly shrinks because of a loss on diversity. With Similarity and \method with $\gamma=0.99$ and $0.999$, the model overfits to a narrow set of most relevant data at around $20$B training tokens. \textit{The persistent long memory of selector slows its adaptivity which can consolidate early-stage preference and repeatedly train on a small set of selected documents.}
With $\gamma=0.1$ and $0$, both downstream and held-out losses decreases smoothly and outperforms the uniform baselines by a large margin. 
The \textit{short-memory} selector adapts to the evolving model preferences immediately, however, it learns little from the history and gives less-accurate assessments on task-relevance. 
With a proper designed memory load at $\gamma=0.9$, \method realizes a good trade-off between adaptivity and task-relevance utility estimate accuracy, achieving the optimal downstream performance.

\textbf{\method recovers from repetition, while static scorer cannot.} 
The design principle refits \method's scorer to the latest model's preference, which endorses the adaptivity of its selection decision. Figure~\ref{fig:repeat_lines} follows the share of documents that are selected for their first-time and ones that are selected for $16$th or more repetitions, which tracks how the selector acquires \textit{diversity} versus \textit{task-relevance}.
With a high memory dependency ($\gamma=0.999$ and $0.99$), \method adapts to acquire more new tokens after the peak of repetitions around $20$B tokens. It aligns with the drastic drop of general FineWeb loss shown in Figure \ref{fig:repeat_lines} (b), where the diversity helps to mitigate the overfitting. With a properly-designed memory at $\gamma=0.9$, \method keeps a good balance between diversity and repetitions along the training, which prevents an overfitting while keeping learning both downstream task and general web at a satisfied speed.
In comparison, the static \textit{Similarity} baseline loses the merit of adaptivity, which leads to a persistent overfitting till the end of training.

\begin{figure*}[t]
\centering
\includegraphics[width=0.95\textwidth]{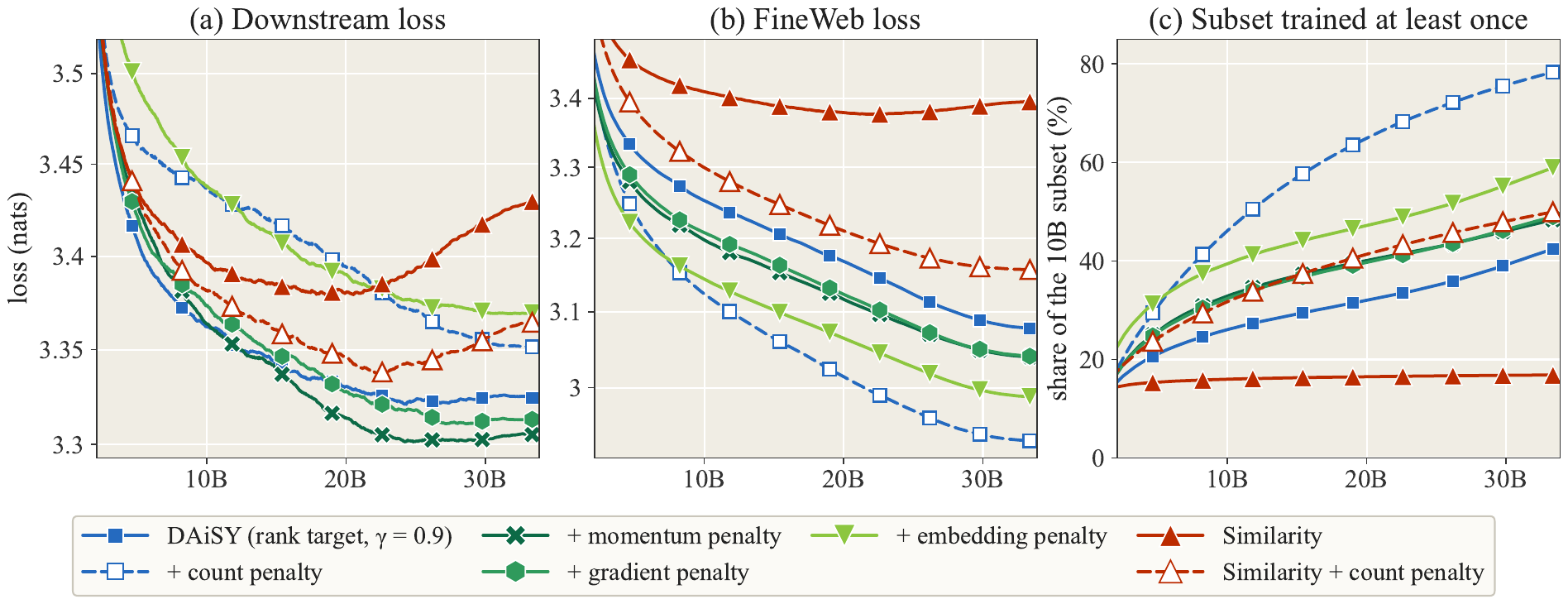}
\caption{\textbf{Redundancy penalties on the 10B subset}. \textbf{(a, b)}~Held-out downstream and FineWeb loss. \textbf{(c)}~Share of the subset trained at least once.}
\label{fig:redundancy}
\vspace{-1.5em}
\end{figure*}
\subsection{How to Design the Redundancy Penalty?}
At the token-boundary, the selector overfits because of \textit{redundancy}: it keeps repeating documents that are relevant to the target while the model has already learned with little marginal values.
To reduce the redundancy during training, we design and evaluate four penalty terms under the token-constrained scenario. 
$z_\gW(x)$ denotes the standardized score among all documents in the current candidate window;
$\beta$ is the hyperparameter that controls the strength of penalty.
\begin{enumerate}[leftmargin=1.4em,itemsep=3pt,topsep=3pt]
\item \textbf{Count penalty.} The training batch is filled with the candidates of highest
\begin{equation}
z_\gW(x)-\beta\log\big(1+c_t(x)\big),
\label{eq:pen_count}
\end{equation}
where $c_t(x)$ is the number of times $x$ has been trained before the current training batch. 
\item \textbf{Embedding penalty.} The micro-batch is filled one document at a time. With $S$ the documents chosen so far, the next one is
\begin{equation}
x^\star=\argmax_{x\in\gW\setminus S}\ z_\gW(x)-\beta\,\big\langle\vu(x),\bar\vu_S\big\rangle,\qquad
\vu(x)=\frac{\vphi(x)}{\|\vphi(x)\|},\qquad \bar\vu_S=\frac{\sum_{x'\in S}|x'|\,\vu(x')}{\sum_{x'\in S}|x'|},
\label{eq:pen_emb}
\end{equation}
until $S$ holds the required micro-batch's tokens. 
The penalty is the \textit{cosine similarity} between a new-coming candidate and the length-reweighted embedding centroid of the documents already chosen.
\item \textbf{Gradient penalty.} Each new-coming document is penalized by the gradient alignment with its siblings in the same batch,
\begin{equation}
y_\beta(x)=z_\gW\big(y(x)\big)-\beta\,z_\gW\big(\rho_{\mathrm{grad}}(x)\big),\qquad\rho_{\mathrm{grad}}(x)=\,\big\langle\mG(x),\mP_t\ \mG(\gB_t\setminus x)\big\rangle
\label{eq:pen_batch}
\end{equation}
where $\mG(x)=\nabla\bar\ell(x)$ is the gradient of the new-coming document, $\mG(\gB_t\setminus x)$ is the gradient of the other documents already chosen in the batch, and $\mP_t$ is the optimizer's preconditioner. We standardize the main utility scores and the penalty term separately because their raw scales differ and drift.

\item \textbf{Momentum penalty.} The penalty term $\rho$ is the alignment between the gradient of the candidate document and the optimizer's first-order momentum, transformed by the optimizer preconditioner $\mP_t$
\begin{equation}
y_\beta(x)=z_\gW\big(y(x)\big)-\beta\,z_\gW\big(\rho(x)\big),\qquad\rho_{\mathrm{mom}}(x)=\big\langle\mP_t\hat\vm_t,\ \nabla\bar\ell(x)\big\rangle,
\label{eq:pen_mom}
\end{equation}
where $\hat\vm_t$ is bias-corrected first moment before step $t$ from the optimizer (i.e. AdamW), which takes an exponential average of the past batch gradients. 
Unlike the in-batch \textit{embedding penalty} and \textit{gradient penalty}, the momentum penalty looks back over several steps rather than at the current batch.
\end{enumerate}
Appendix~\ref{app:redundancy} gives further details of the setups and compelete results.

\paragraph{Gradient-based penalty with long-horizon is the most effective.} Figure~\ref{fig:redundancy} and Table~\ref{tab:redundancy} compare the four penalties terms. \textit{Count penalty} applies a universal static penalty on all documents, which increases the coverage of documents to $80\%$ while not helping with the downstream task. 
Similarly, \textit{embedding penalty} improves the coverage to $60\%$ but fails to improve upon the downstream loss. It indicates that acquiring more representation-level diversity or point-less document coverage is NOT ideal for downstream improvement. 

In comparison, \method with either \textit{gradient penalty} or \textit{momentum penalty} increase the coverage by $+10\%$ while successfully pushing both downstream task and general FineWeb learning. With a long-horizon view, momentum penalty is the most effective method with the optimal final downstream loss and held-out FineWeb loss. Besides, since the \textit{momentum penalty} makes use of the on-the-shelf optimizer's first-order momentum, it does not incur any extra computation or memory overhead. Therefore, we consider the \textit{momentum penalty} as an ideal penalty to apply in practice.


\section{Conclusion}
We introduce \method, a lightweight and powerful online data selection pipeline. We demonstrate its effectiveness and efficiency under English pretraining and multilingual pretraining scenarios, with various steering target designs. The improvements from \method can consistently survive the instruction-tuning phase. We also consolidate the strength of an \textit{online} data selector at token boundaries to dynamically keep the trade-off between task-relevance and diversity.

\subsection{Limitations and Future Works.}
Because of the large compute costs from pretraining jobs and limited computation resources, we only train each run once using a single random seed. However, we believe the benefits from \method are significant enough above a random-seed variance range. 
Also, according to our observation, the benefit from \method has not been capped by the candidate window $\lambda=64$, but we have not explored above it.

\textbf{Future works.} \method demonstrates the promising power of a linear data selector backed with a static embedding model. We also validate its effectiveness under noisy and highly heterogeneous space on multilingual pretraining, where the six languages come from various language families with very different vocabularies. For future works, we expect to apply \method on more heterogeneous and challenging task space including world models and physical AI with information coming from multiple modalities.

\newpage
\clearpage
\bibliography{main}
\bibliographystyle{abbrvnat}

\newpage
\appendix
\section{Model Configurations}
\label{app:sec:setup}
Table~\ref{tab:modelconfig} gives the configuration of the three models that we train: the English 124M and 720M models with GPT2 tokenizer, and the 124M model of the multilingual runs (Section~\ref{sec:multilingual}).

\begin{table}[h]
\centering
\caption{\textbf{Model and training configuration.}}
\label{tab:modelconfig}
\footnotesize
\begin{tabular}{lccc}
\toprule
& 124M (English) & 720M (English) & 124M (multilingual) \\
\midrule
\multicolumn{4}{l}{\emph{Architecture}} \\
Layers & 12 & 12 & 12 \\
Width & 768 & 2,048 & 768 \\
Attention heads (query and key/value) & 12 & 16 & 12 \\
Head dimension & 64 & 128 & 64 \\
SwiGLU hidden width & 2,048 & 5,632 & 2,048 \\
Context length (tokens) & 2,048 & 2,048 & 2,048 \\
Vocabulary & 50,304 (GPT-2) & 50,304 (GPT-2) & 200,064 (Apertus) \\
\midrule
\multicolumn{4}{l}{\emph{Parameters}} \\
Transformer blocks and final norm & 85.0M & 616.6M & 85.0M \\
Embedding (tied with the output layer) & 38.6M & 103.0M & 153.6M \\
Total & 123.6M & 719.6M & 238.6M \\
\midrule
\multicolumn{4}{l}{\emph{Training}} \\
GPUs (GH200) & 4 & 16 & 4 \\
Rows of 2,048 tokens per GPU and step & 16 & 16 & 16 \\
Tokens per step & 131,072 & 524,288 & 131,072 \\
Steps & 256,000 & 64,000 & 256,000 \\
Training tokens & 33.6B & 33.6B & 33.6B \\
\bottomrule
\end{tabular}
\end{table}

\newpage
\clearpage

\section{Implementation}
\label{app:impl}
The following pseudo-code gives the core implementation of \method.
\begin{lstlisting}
# ---- target direction, every R steps ----------------------------------------------------------
g = grad(token_mean_loss(model, target_rows))                       # one target backward, no .grad writes
v_hat = {n: opt.exp_avg_sq[n] / (1 - beta2 ** t) for n in linear_names}   # bias-corrected 2nd moment
V = {n: g[n] / (v_hat[n].sqrt() + opt.eps) for n in linear_names}         # Adam preconditioner

# ---- label gate on every Linear (same backward as training), on every L-th step ----------------
def bwd_hook(mod, grad_in, grad_out):                                # passes grad_out through unchanged
    a, d = acts.pop(mod).float(), grad_out[0].detach().flatten(0, -2).float()
    labels.index_add_(0, doc_ids, ((d @ V[mod]) * a).sum(-1))        # delta^T V a, summed per document
# y[x] = labels[x] * N_batch_tokens / len(x): utility of the document's mean token loss

# ---- online ridge, one process, every labeled step from t0 on; solve every U steps -------------------
r = norm_ppf((rank(y) - 0.5) / len(y))                              # Gaussianized ranks within the step
X = cat([phi[docs], ones]);  A = gamma * A + X.T @ X;  b = gamma * b + X.T @ r
if t % U == 0:                                                      # the bias is not regularized
    reg = lam * A[:-1, :-1].trace() / (len(A) - 1) * diag([1] * (len(A) - 1) + [0])
    w = solve(A + reg, b);  broadcast(w)

# ---- selection, per micro-batch ----------------------------------------------------------------
cand = stream.next_until(tokens >= lam_window * micro_batch_tokens)
explore = uniform_take(cand, eps * micro_batch_tokens)
chosen = explore + top_by(cand - explore, key=lambda x: w @ cat([phi[x], 1]))   # fill, cut the last one
\end{lstlisting}

\newpage
\clearpage
\section{Cost Analysis}
\label{app:costmethods}
This appendix compares the runs on the efficiency axes, including GPU-hour, memory and FLOPS costs. 
The rest of the section covers detailed cost breakdown on each method we implemented, including \method, \textit{Similarity}, OPUS and MATES.
Table~\ref{tab:costmethods} gives the measured cost of every method at 124M and 720M. 
\emph{Step time} is the steady-state mean over the steps recorded across the entire 33.6B token run; 
\emph{Peak memory} is the largest allocated memory on one H200 GPU. 
\begin{table}[ht!]
\centering
\caption{\textbf{Cost of every method on Nvidia-H200 GPU.} Every entry is measured per GPU. \textbf{\textit{GPU-hours}} reports the entire GPU time cost on 33.6B training tokens, 256k steps at 124M and 64k at 720M; 
\textit{\textbf{$\times$standard}} reports step time over standard uniform run.}
\vspace{0.2em}
\label{tab:costmethods}
\footnotesize
\begin{tabular}{lrrrrrr}
\toprule
& \multicolumn{4}{c}{124M} & \multicolumn{2}{c}{720M} \\
\cmidrule(lr){2-5}\cmidrule(lr){6-7}
Arm & s/step & $\times$standard & GPU-h & Peak GiB & s/step & GPU-h \\
\midrule
Uniform & 0.1012 & 1.00 & 28.8 & 16.2 & 0.4099 & 116.6 \\
Similarity $\lambda=2$ & 0.1126 & 1.11 & 32.0 & 16.2 & -- & -- \\
Similarity $\lambda=4$ & 0.1153 & 1.14 & 32.8 & 16.2 & 0.4227 & 120.2 \\
Similarity $\lambda=8$ & 0.1184 & 1.17 & 33.7 & 16.2 & -- & -- \\
Similarity $\lambda=16$ & 0.1247 & 1.23 & 35.5 & 16.2 & 0.4286 & 121.9 \\
Similarity $\lambda=32$ & 0.1269 & 1.25 & 36.1 & 16.2 & -- & -- \\
Similarity $\lambda=64$ & 0.1528 & 1.51 & 43.5 & 16.2 & 0.4453 & 126.6 \\
DSIR & 0.1219 & 1.20 & 34.7 & 16.2 & -- & -- \\
\method $\lambda=2$ & 0.1147 & 1.13 & 32.6 & 22.2 & -- & -- \\
\method $\lambda=4$ & 0.1183 & 1.17 & 33.7 & 21.2 & 0.4342 & 123.5 \\
\method $\lambda=8$ & 0.1168 & 1.15 & 33.2 & 22.2 & -- & -- \\
\method $\lambda=16$ & 0.1208 & 1.19 & 34.4 & 22.2 & 0.4542 & 129.2 \\
\method $\lambda=32$ & 0.1347 & 1.33 & 38.3 & 21.6 & -- & -- \\
\method $\lambda=64$ & 0.1527 & 1.51 & 43.4 & 21.1 & 0.4599 & 130.8 \\
OPUS (documents-wise) & 0.7353 & 7.27 & $\approx$209$^{a}$ & 51.6 & -- & -- \\
OPUS (sequences-wise) & 0.4402 & 4.35 & $\approx$125$^{a}$ & 17.2 & -- & -- \\
MATES & 0.1004$^{b}$ & 2.31$^{c}$ & 66.5$^{c}$ & 27.7 & -- & -- \\
MATES, first boundary at step 2,500 & 0.1059$^{b}$ & 2.67$^{c}$ & 76.8$^{c}$ & -- & -- & -- \\
\bottomrule
\end{tabular}
\\[2pt]{\footnotesize $^{a}$\,Extrapolated from the step time. The runs stopped earlier: OPUS (sequence-wise) after 65,200 (31.9 GPU-hours); OPUS (document-wise) at step 113,600 (92.6 GPU-hours); \\$^{b}$\,Training step time between stage boundaries; each boundary adds the selector tuning and selection costs ( Appendix~\ref{app:cost:mates}). \\$^{c}$\,Measured over the whole \textit{selection-and-training} schedule, including the boundary budget.}
\end{table}

\newpage
\clearpage
\subsection{Performance at Equal Compute}
\label{sec:cost}
\begin{figure}[ht!]
\centering
\includegraphics[width=0.7\textwidth]{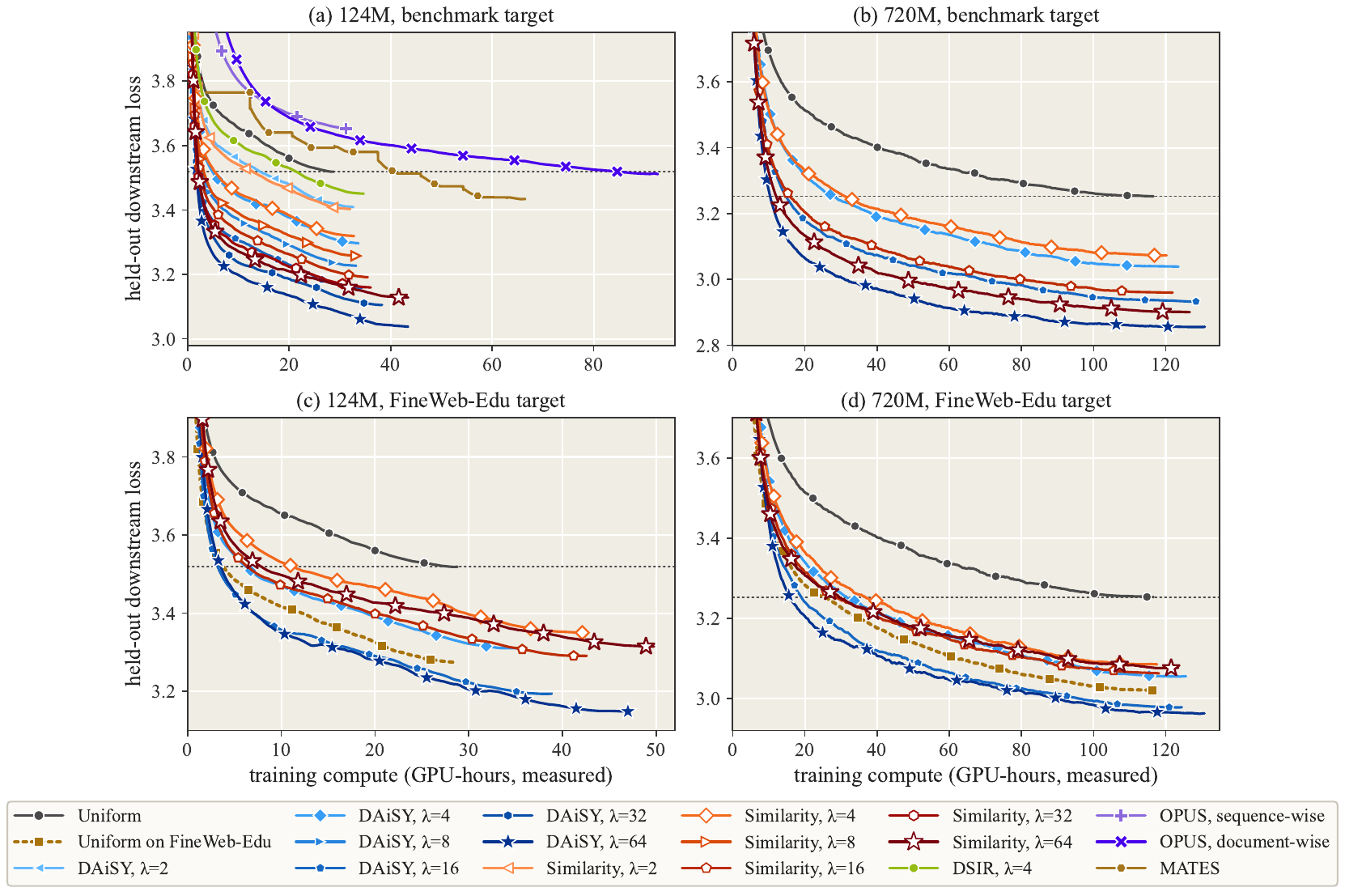}
\caption{\textbf{Held-out downstream loss against measured GPU-hours}. \textbf{(a, b)}~Benchmark target at 124M and 720M: the runs of Figure~\ref{fig:task_orient}, including OPUS runs on sequence-wise selection. \textbf{(c, d)}~FineWeb-Edu target at 124M and 720M: the runs of Figure~\ref{fig:edu_efficiency}. 
The dotted horizontal line marks the final loss of a standard uniform training baseline. The curve of MATES, shown in (a) only, runs flat while a stage boundary training the selector and selecting data.}
\label{fig:efficiency_gpuh}
\end{figure}

Figure~\ref{fig:efficiency_gpuh} provides the downstream loss curves according to GPU hours target at both scales. 

\subsection{Summary of the overhead}
\label{sec:costmodel}
\method adds four operations to training: 
the label gates on every $L$-th step, one target refresh pass every $R$ steps, one dot product for scoring per candidate document, and one $(d{+}1)\times(d{+}1)$ linear ridge regression solve every $U$ steps. 
None of them is a forward or backward pass upon non-selected candidates, so the overhead of selection is negligible comparing to LLM training step. 
We report measurements at 124M with $\lambda=4$, $L=4$, $U=10$ and $R=200$; Table~\ref{tab:costmethods} lists every arm, and the following subsections give the full accounting.

\paragraph{Overhead of \method.} In FLOPs, the label gates add $\approx5.0\%$ to training, and the target gradient refresh pass adds $\approx0.6\%$. Scoring the candidates and solving the ridge regression are negligible ($<10^{-4}$\%), which gives a total overhead of about $\approx5.6\%$.
In time, the overhead is larger from reading the embeddings of $\lambda\times$ more documents occupies some data-loader threads, which costs about $+3\%$ at 124M. But since this time cost is independent to FLOPs or model scale, it can be negligible when scaling up to larger models ($\approx 0.3\%$ for a 1B model).

\paragraph{Comparison with model-aware selection baselines.} The baselines that score candidates with the model cost far more. OPU\citep{wang2026opus} scores only the first 512 tokens per document (or per sequence-chunk), while still leads to $7.3\times$ time costs on documents-wise ($4.35\times$ on sequence-wise). MATES spends 37.0 out of its 66.5 GPU-hours (56\%) in tuning the BERT selector and scoring documents at stage boundaries, which amounts to $2.3\times$ the cost of plain training over the schedule.

\subsection{\method FLOP decomposition at 124M}
\label{app:pilotcost}
\label{app:cost}
\paragraph{Conventions.} We count a matrix product of an $n\times a$ by an $a\times b$ matrix as $2nab$ FLOPs, a backward pass as twice the forward pass, and causal self-attention as $2\cdot2\cdot d\cdot T/2$ FLOPs per token and layer in the forward pass. All counts are per optimizer step of $2048\times 64 = 131,072$ tokens, averaged over the steps after step 2500. Evaluation passes are measurement and not method cost.

\paragraph{Training.} With $L=12$ blocks, $d=768$, SwiGLU width 2048, vocabulary 50,304 and context 2048, the linear layers inside the blocks have 84.9M weights and the readout has 38.6M. A forward pass costs $2.85\times10^{8}$ FLOPs per token and a training step $8.55\times10^{8}$ FLOPs per token, that is, $1.12\times10^{14}$ FLOPs per step.

\paragraph{Labels.} Consider a linear layer $d_{\text{in}}\to d_{\text{out}}$. Each training step, each token costs about $6d_{\text{in}}d_{\text{out}}$ FLOPs for the forward-backward pass. 
The gate of Eq.~\ref{eq:ghost} computes $\mV\va$ and its inner product with $\vdelta$, which costs $2d_{\text{in}}d_{\text{out}}$ FLOPs, one third of the training cost on the linear layers on a labeled step. Nothing is computed on non-linear layers like attention layers. 
Over all the trainable linear layers, the gates cost $1.70\times10^{8}$ FLOPs per token and target, which is $19.9\%$ of a training step, i.e. $4.97\%$ overhead amortized by label collection frequency $L=4$. 

\paragraph{Target refresh and candidate scoring costs.} The target gradient costs $1.31\times10^{14}$ FLOPs per refresh amortized by $R=200$ steps (0.59\% of a training step). Each candidate scoring costs one 769-dimensional (768 for embeddings and 1 for bias) dot product, which is negligible in FLOPs. Embedding loading costs time but with no extra FLOPs.

\subsection{Similarity scorer}
\label{app:cost:sim}
The similarity scorer reads the precomputed embeddings of $\lambda\times$ candidates than it trains on, scores each with one 769-dimensional dot product, and packs the chosen documents. 

\subsection{DSIR}
\label{app:cost:dsir}
The DSIR score of a document is a sum over its hashed n-grams, so it is precomputed once before training on CPUs. During training, the selector looks up the scores and packs the chosen documents as similarity does. 

\subsection{OPUS}
\label{app:cost:optiselect}
We run OPUS~\citep{wang2026opus} in two versions that differ in the unit of selection: (1) \textit{document-wise}: selecting in document level, where at most one document can be truncated; (2) \textit{sequence-wise}: selecting in 2048 sequence-chunk, where every document can be truncated or patched together. Figure \ref{fig:task_orient} compares with the OPUS with \textit{document-wise} selection, which yields better token-efficiency. 

We follow the default setup in \citep{wang2026opus} and only score each candidate on its first 512 tokens out of 2048 sequence chunk (\textit{sequence-wise}) or each document (\textit{document-wise}). 
The scored tokens take the full forward-backward pass to get the exact gradient-based utility scores, yielding $\lambda \times 100\%$ extra FLOPS of a standard training step. 
Additionally, a low-rank gradient sketch and gradient product also costs extra overheads. Overall, the profiled step takes 0.446\,s, $4.5\times$ of the costs on a standard training step.

\subsection{MATES}
\label{app:cost:mates}
MATES~\citep{yu2024mates} trains in stages, and at every stage boundary it 
(i) \textit{probes held-out documents}, taking one training step on each document with a copy of the current model and optimizer state, then computing the reference loss over the five targets (153,600 tokens in total); 
(ii) fine-tunes a BERT-base data influence model on the probes; and (iii) scores a fresh pool of candidates that holds $5\times$ the training tokens of the stage and keeps 20\% of them by Gumbel-top-$k$. 
We run it at 124M with stages of 32,000 steps. We apply 80,000 probes at the first boundary and 20,000 later following the default setup in \citep{yu2024mates}, which explains the longer time cost on the first stage in Figure \ref{fig:efficiency_gpuh}.
Stage~1 before the first boundary is equivalent to the plain uniform run. 

\paragraph{MATES with first boundary at step 32000.} \citet{yu2024mates} set the first boundary at step 32k. Across the full run, the boundaries at steps 32k, 64k and 96k cost 8.67, 4.63 and 4.61 GPU-hours, each scoring $\lambda=5\times$ of tokens required for the next stage and kept 20.0\% of the tokens of its candidates. 

\paragraph{MATES with first boundary at step 2,500.} We also run the MATES variant that selects from step 2,500, where the scorer of \method starts. The boundaries at steps 2.5k, 34.5k, 66.5k cost 9.01, 4.72, 4.71 and 4.71 GPU-hours respectively. 
Continued to the full 256k run, it consumes 76.80 GPU-hours in total: 33.84 for training on selected data and 42.94 at eight boundaries.

\newpage
\clearpage
\section{Gaussianized Ranks}
\begin{figure}[ht!]
\centering
\includegraphics[width=\textwidth]{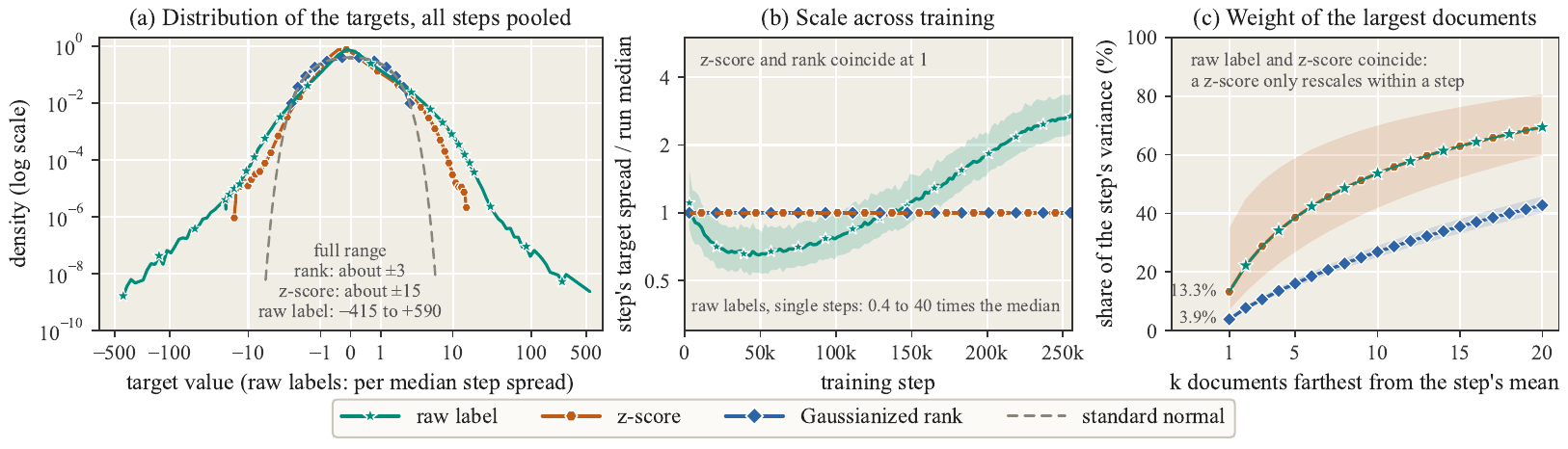}
\caption{\textbf{Raw labels, z-scores and Gaussianized ranks as regression targets} (124M run with 25\% exploration). \textbf{(a)}~Density of each target over all 13.1M labeled documents, with raw labels in units of the median step's standard deviation. Ranks follow the standard normal (dashed), whereas z-scores and raw labels have heavy tails. \textbf{(b)}~Standard deviation of a step's targets relative to the run's median (line: median; shaded: 5th to 95th percentile over steps). \textbf{(c)}~Share of a step's variance that its $k$ most extreme documents carry; raw labels and z-scores coincide.}
\label{fig:targets}
\end{figure}

In Figure \ref{fig:targets}, we present the distributions of each labels (raw utility scores, z-score with linear standardization, and Gaussianized ranks), and compare their scale and variance. 

\newpage
\clearpage
\section{Additional Details for Task-Oriented Pretraining}
We present detailed statistics and curves for experiments in task-oriented pretraining section (\S\ref{sec:benchmark-target}).
Table \ref{tab:generality} presents the held-out downstream loss on targeted benchmarks and held-out FineWeb loss, which shows that the downstream performance gain costs a general learning on web data.

\begin{table}[ht!]
\centering
\caption{\textbf{Held-out FineWeb loss of the runs of Table~\ref{tab:task_orient}} $\Delta$ denotes the distance against uniform training run, lower means improvements while higher means degradation.}
\label{tab:generality}
\footnotesize
\setlength{\tabcolsep}{5pt}
\begin{tabular}{llcccc}
\toprule
& & \multicolumn{2}{c}{Downstream loss} & \multicolumn{2}{c}{FineWeb loss} \\
\cmidrule(lr){3-4}\cmidrule(lr){5-6}
$\lambda$ & Scorer & Loss & $\Delta$ & Loss & $\Delta$  \\
\midrule
\multicolumn{6}{l}{\emph{124M}} \\
\multicolumn{2}{l}{Uniform} & 3.521 & -- & 2.887 & --  \\
\midrule
5 & MATES & 3.435 & $-0.086$ & 2.887 & $+0.000$ \\
\midrule
\multirow{3}{*}{4} & DSIR & 3.452 & $-0.069$ & 2.935 & $+0.048$ \\
& Similarity & 3.321 & $-0.200$ & 3.000 & $+0.113$ \\
 & \cellcolor{daisyrow}\method & \cellcolor{daisyrow}3.299 & \cellcolor{daisyrow}$-0.222$ & \cellcolor{daisyrow}3.001 & \cellcolor{daisyrow}$+0.114$ \\
\midrule
\multirow{2}{*}{16} & Similarity & 3.194 & $-0.327$ & 3.172 & $+0.285$ \\
 & \cellcolor{daisyrow}\method & \cellcolor{daisyrow}3.160 & \cellcolor{daisyrow}$-0.361$ & \cellcolor{daisyrow}3.163 & \cellcolor{daisyrow}$+0.276$ \\
\midrule
\multirow{2}{*}{64} & Similarity & 3.129 & $-0.392$ & 3.317 & $+0.430$ \\
 & \cellcolor{daisyrow}\method & \cellcolor{daisyrow}3.040 & \cellcolor{daisyrow}$-0.480$ & \cellcolor{daisyrow}3.278 & \cellcolor{daisyrow}$+0.391$ \\
\midrule
\multicolumn{6}{l}{\emph{720M}} \\
\multicolumn{2}{l}{Uniform} & 3.255 & -- & 2.590 & --  \\
\midrule
\multirow{2}{*}{4} & Similarity & 3.075 & $-0.180$ & 2.708 & $+0.118$\\
 & \cellcolor{daisyrow}\method & \cellcolor{daisyrow}3.040 & \cellcolor{daisyrow}$-0.214$ & \cellcolor{daisyrow}2.708 & \cellcolor{daisyrow}$+0.118$ \\
\midrule
\multirow{2}{*}{16} & Similarity & 2.961 & $-0.294$ & 2.873 & $+0.283$ \\
 & \cellcolor{daisyrow}\method & \cellcolor{daisyrow}2.935 & \cellcolor{daisyrow}$-0.319$ & \cellcolor{daisyrow}2.875 & \cellcolor{daisyrow}$+0.285$ \\
\midrule
\multirow{2}{*}{64} & Similarity & 2.902 & $-0.353$ & 3.015 & $+0.425$\\
 & \cellcolor{daisyrow}\method & \cellcolor{daisyrow}2.856 & \cellcolor{daisyrow}$-0.399$ & \cellcolor{daisyrow}2.989 & \cellcolor{daisyrow}$+0.399$ \\
\bottomrule
\end{tabular}
\end{table}

\begin{figure}[ht!]
\centering
\includegraphics[width=0.8\textwidth]{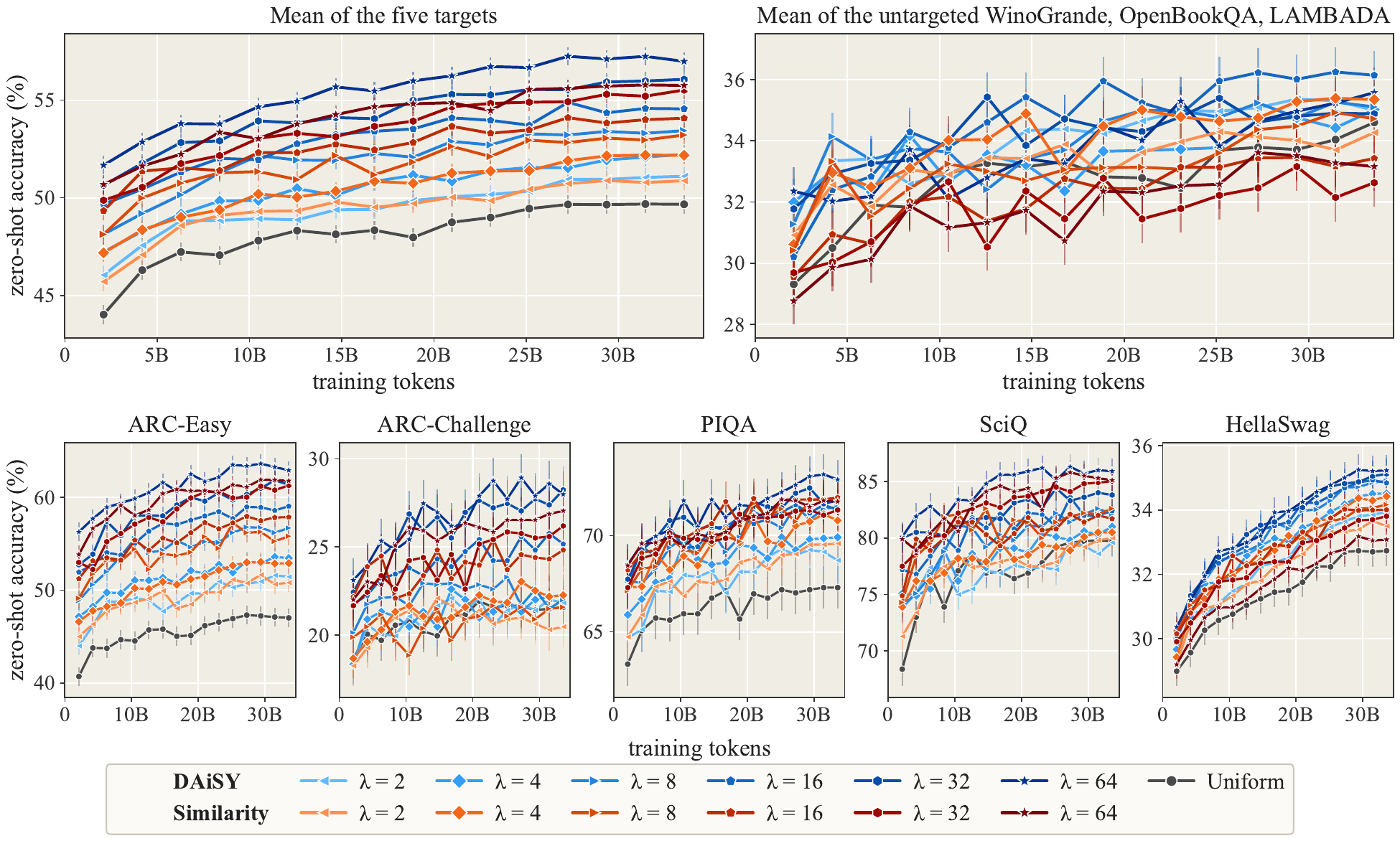}
\caption{\textbf{Zero-shot accuracy curves with task-oriented pretraining at 124M}. Top: mean over the five targets (left) and over the three untargeted benchmarks (right). Bottom: each target.}
\label{fig:lambdaacc}
\end{figure}

\begin{figure}[ht!]
\centering
\includegraphics[width=0.8\textwidth]{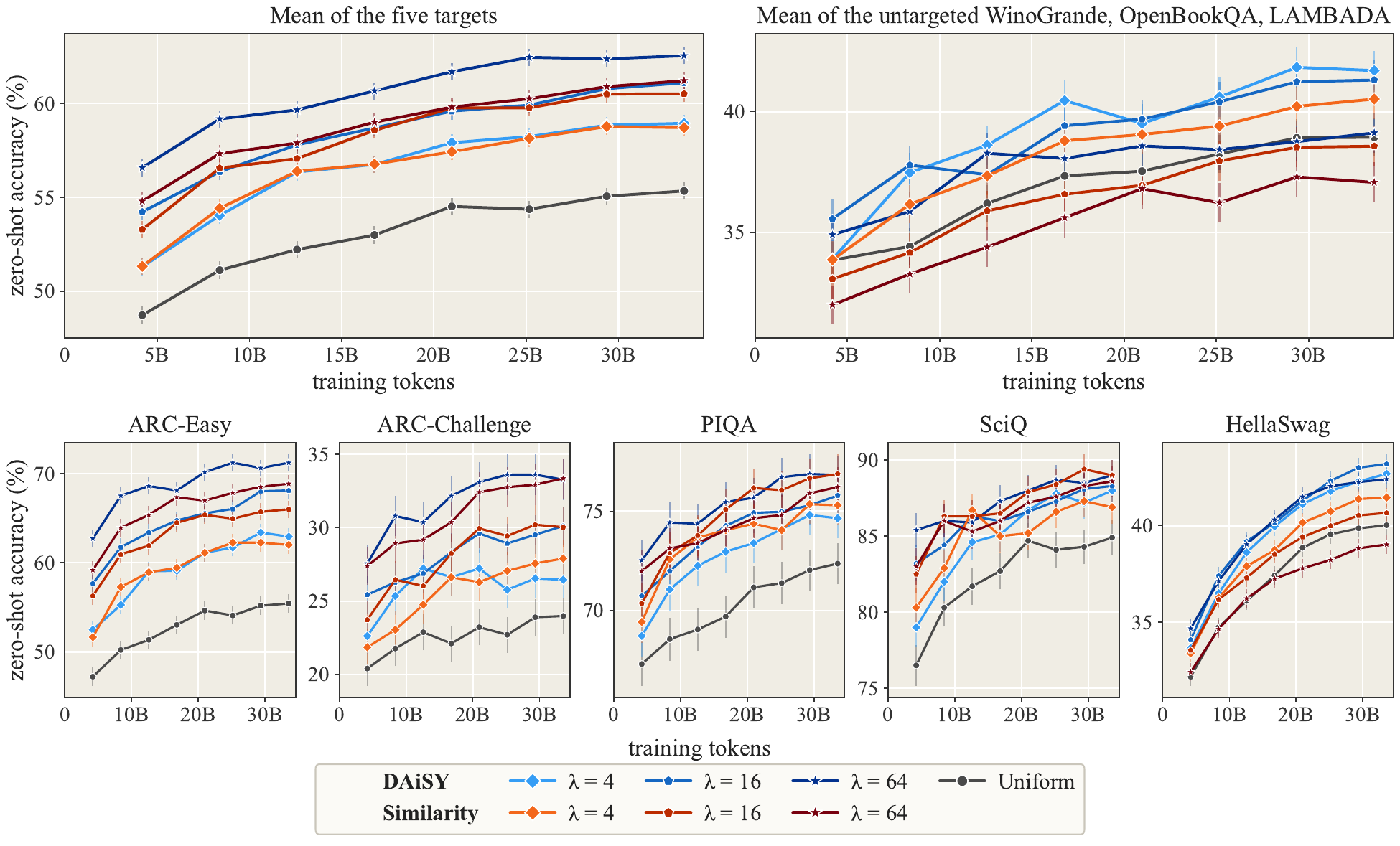}
\caption{\textbf{Zero-shot accuracy curves with task-oriented pretraining at 720M}. Top: mean over the five targets (left) and over the three untargeted benchmarks (right). Bottom: each target.}
\label{fig:lambda720}
\end{figure}

\newpage
\clearpage
\section{Additional Details on Web-Targeted Pretraining}
\label{app:webtargets}
We describe the detailed procedure on how we conduct topic clustering and construct the FineWeb and HQ FineWeb-Edu targets (\S\ref{app:web-setup}).
We also include detailed statistics and results of the runs of Section~\ref{sec:webtargets} in \ref{app:web_results}, with a topic-wise analysis at 720M.

\subsection{Setup}
\label{app:web-setup}
\paragraph{FineWeb topic target}
We run spherical $k$-means with 16 clusters on the L2-normalized embeddings $\vphi$ of a sample of 2M documents (k-means++ initialization, 100 Lloyd iterations) and assign every document to its nearest centroid. 
Each clusters hold 3.8M--9.2M documents.
We reserve 1,936 held-out documents of each topic to form the target. They are strictly exclude from evaluation and training corpus, so the LLM has never seen them during training. 
Reserve documents whose text or URL hash equals that of an evaluated document are removed for \textit{decontamination}. 
The target averages the gradients across topics, so it weighs the 16 topics equally. 

\paragraph{FineWeb-Edu target}
We use FineWeb-Edu~\citep{penedo2024fineweb}, which is a high-quality subset from FineWeb. An where educational-quality classifier assign a quality score for each FineWeb documents from 0 to 5. 
We keep the documents with a score at least 4, leaving 33,256 documents. 
We then remove 2,191 documents whose text and 642 whose URL equals that of documents in held-out FineWeb evaluation set, leaving 30,423 in the end.
The FineWeb-Edu dataset keeps scores of at least 3, so we train the FineWeb-Edu baseline (Figure \ref{fig:edu_efficiency}) on this with uniform sampling. We choose 1024 documents from them to form the held-out FineWeb-Edu set, which is strictly excluded from the HQ FineWeb-Edu target.

\subsection{Results}
\label{app:web_results}
\begin{figure}[ht!]
\centering
\includegraphics[width=0.76\textwidth]{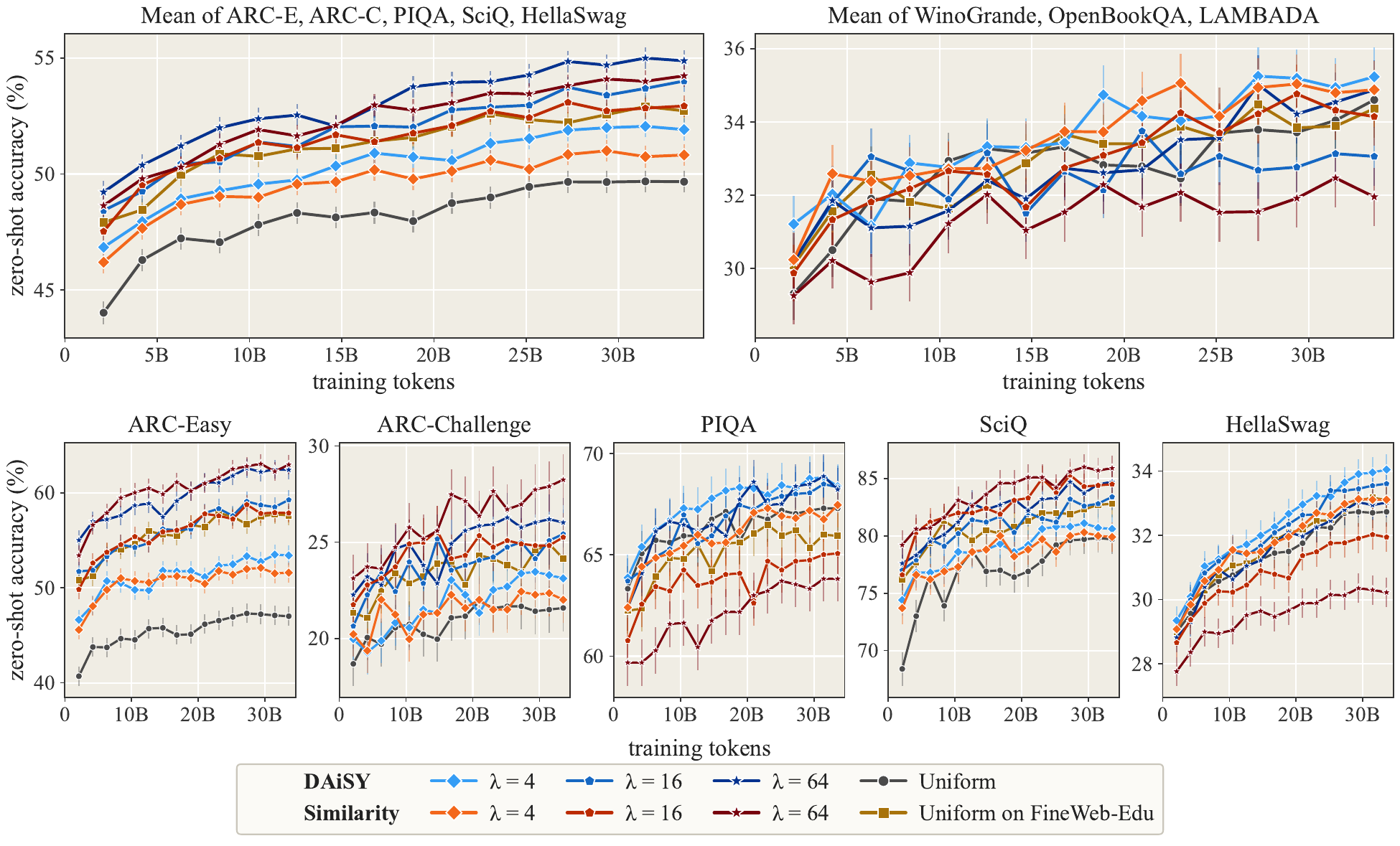}
\caption{\textbf{Zero-shot accuracy curves with task-oriented pretraining at 124M}. Top: mean over the five targets (left) and over the three untargeted benchmarks (right). Bottom: each target.}
\label{fig:eduacc124}
\end{figure}

\begin{figure}[ht!]
\centering
\includegraphics[width=0.76\textwidth]{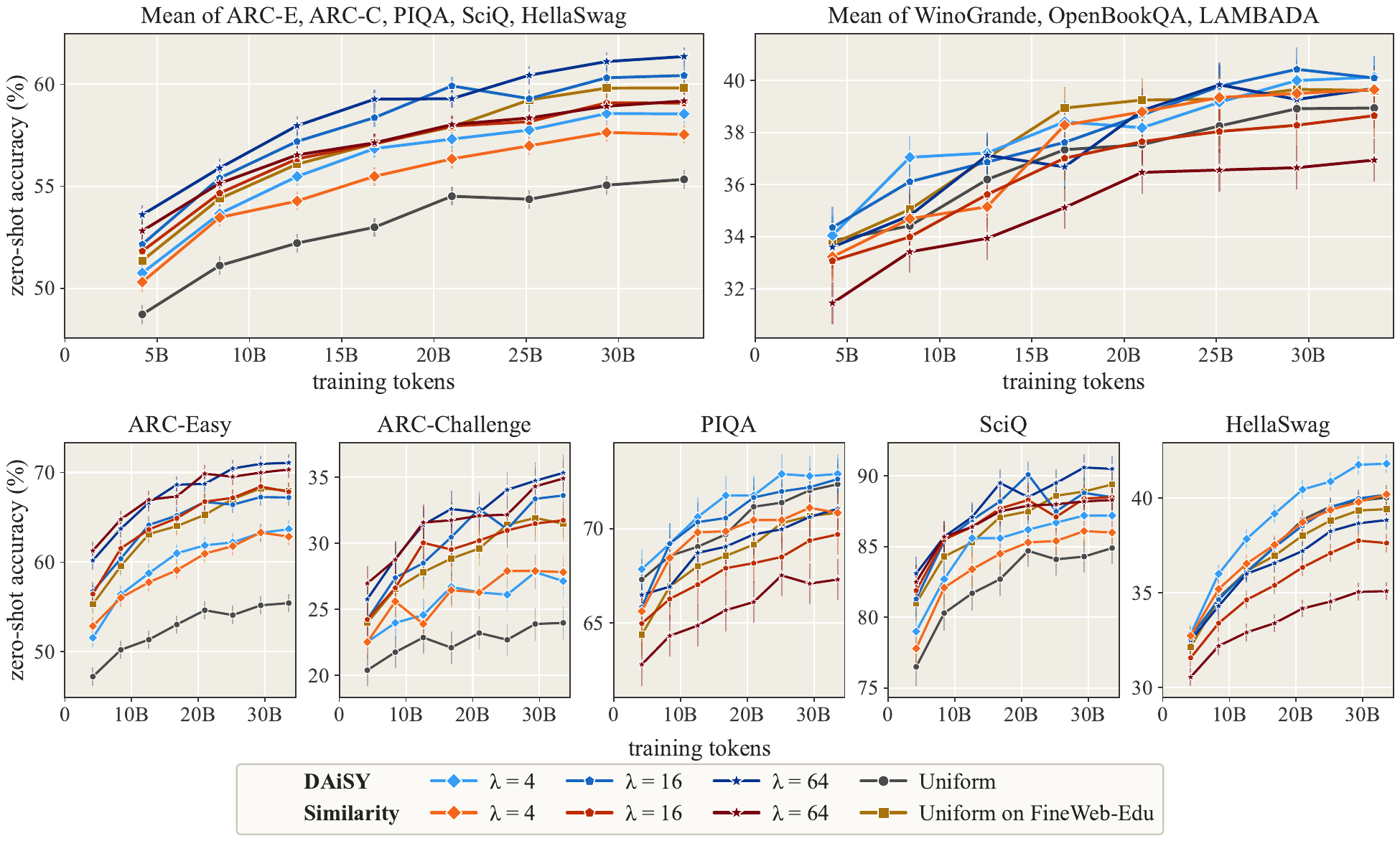}
\caption{\textbf{Zero-shot accuracy curves with task-oriented pretraining at 720M}. Top: mean over the five targets (left) and over the three untargeted benchmarks (right). Bottom: each target.}
\label{fig:eduacc720}
\end{figure}

\begin{figure*}[ht]
\centering
\includegraphics[width=0.7\linewidth]{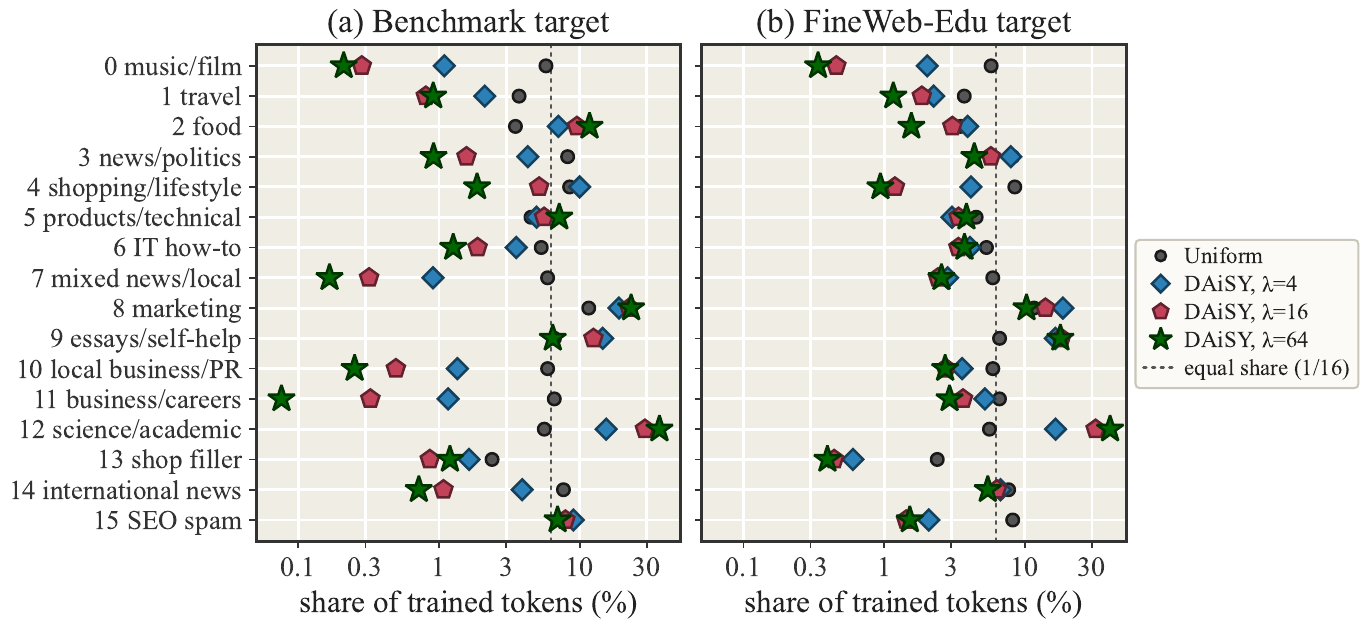}
\caption{\textbf{Topic composition of the trained documents at 720M}, as Figure~\ref{fig:selection_topics}, for the benchmark target and the FineWeb-Edu target.}
\label{fig:selection_topics720}
\end{figure*}

We present per-task accuracy curves at 124M and 720M in Figure \ref{fig:eduacc124} and Figure \ref{fig:eduacc720}. The selected topic compositions in Figure \ref{fig:selection_topics} and Figure \ref{fig:selection_topics720}.

\begin{figure*}[ht]
\centering
\includegraphics[width=0.7\linewidth]{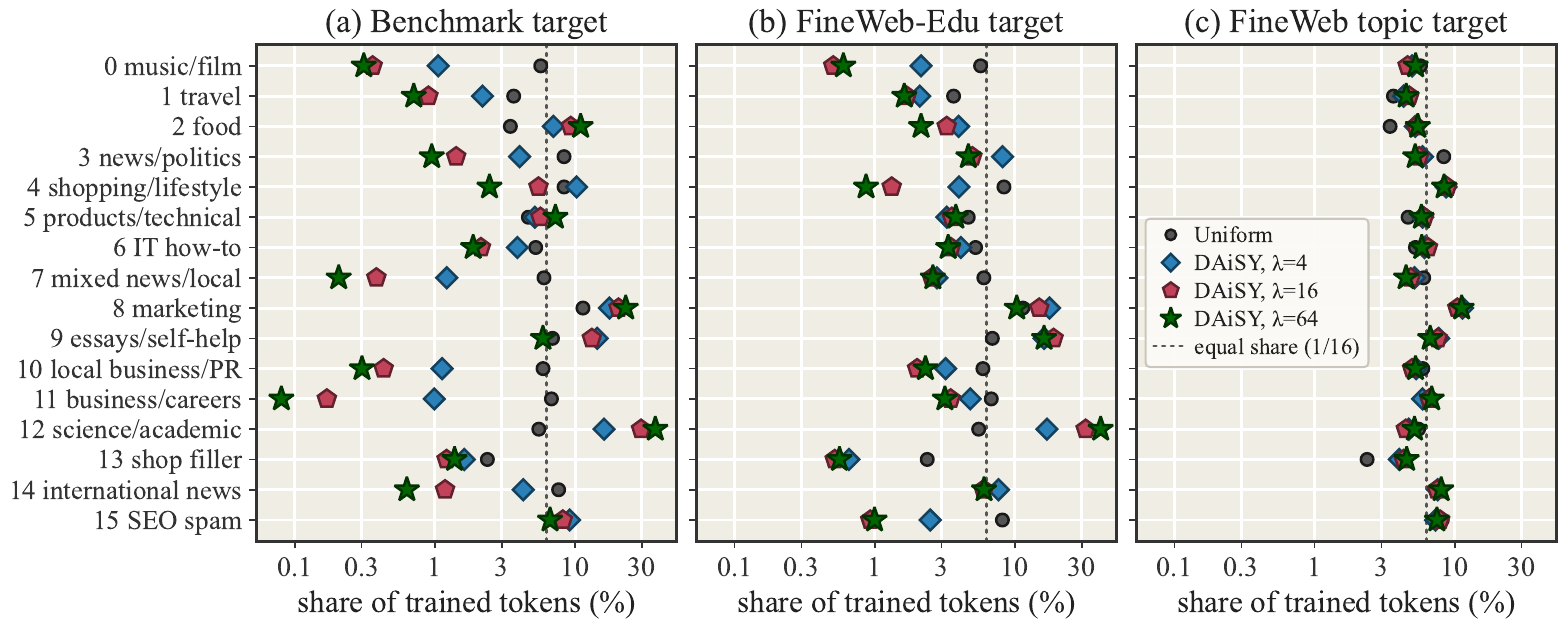}
\caption{\textbf{Topic composition of the trained documents at 124M}.}
\end{figure*}

\newpage
\clearpage
\section{Additional Details on Instruction Tuning}
\label{app:sft}
This appendix gives the details configurations of the five-epoch instruction tuning of Section~\ref{sec:sft} in \S\ref{app:sec:instruct-setup}. We also present additional evaluation on IFEval and analysis on the forgetting during instruction tuning in \S\ref{app:sec:instruct-results}.
\subsection{Setup.}
\label{app:sec:instruct-setup}
\paragraph{Chat template.} The zero-shot evaluation presents every question as plain text, whereas the tuning data is in chat format. We therefore also render each question as a user turn and score each answer as the assistant's reply, as in training. We also leave out LAMBADA as it is a next-token prediction task, not a reasoning task.

\paragraph{Schedule.} The recipe is that of Appendix~\ref{app:sft} except for its length and schedule. 
The run takes 31,390 steps in total for five epochs, warms up over 3\% of them and then keeps the learning rate at $10^{-4}$. Before the end of each of the five epochs, a copy of the model is annealed linearly to zero over the last $20\%$ epochs (1,256 steps) and saved, and the run continues from the state before the copy. The model after $k$ epochs is therefore properly annealed. The document order is re-shuffled in each epoch.

\subsection{Results}
\label{app:sec:instruct-results}
\begin{figure}[ht]
\centering
\includegraphics[width=0.92\textwidth]{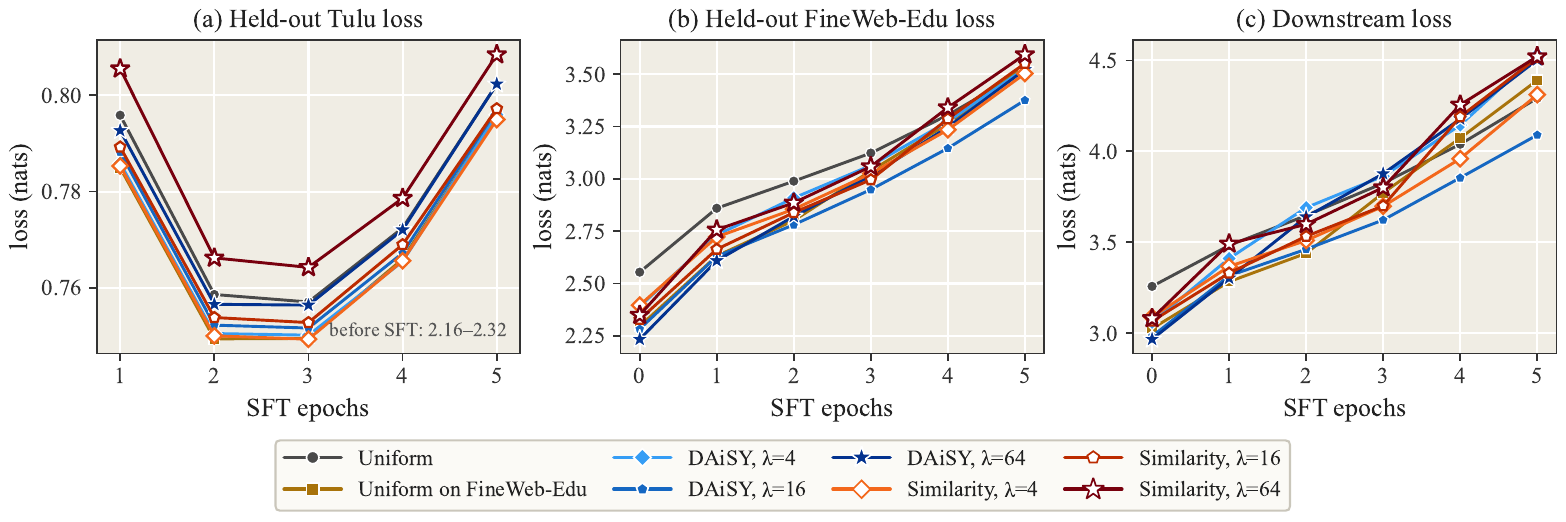}
\caption{\textbf{Five epochs of instruction tuning at 720M.} (a) Held-out Tulu loss; (b) held-out FineWeb-Edu loss; (c) downstream loss.}
\label{fig:sft_forgetting}
\end{figure}

\paragraph{Forgetting.} Figure~\ref{fig:sft_forgetting} shows the three losses. The Tulu loss is lowest to 0.749--0.764 after two or three epochs and rises to 0.795--0.808 after five epoches. The FineWeb-Edu loss and the downstream loss rise with every epoch. Longer tuning therefore does not always improve reasoning abilities, while leading to a consistent forgetting on the general web data from pretraining.

\begin{figure}[ht]
\centering
\includegraphics[width=0.5\textwidth]{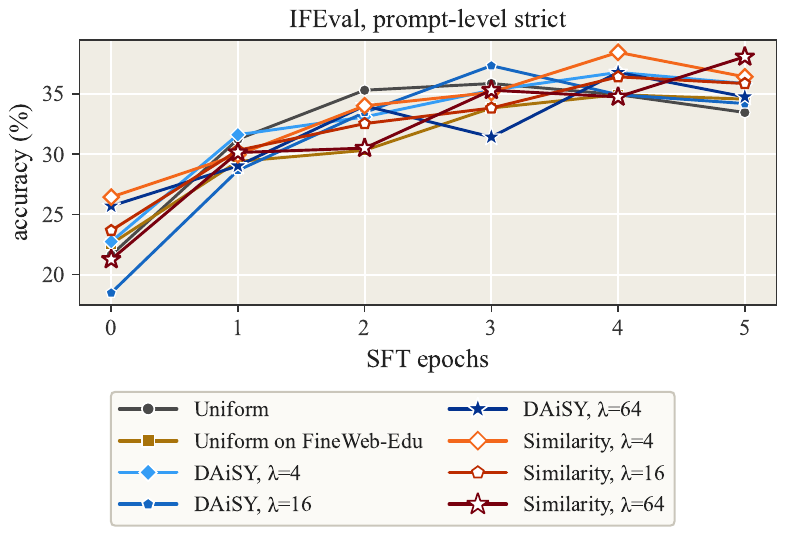}
\caption{\textbf{IFEval after $k$ epochs of instruction tuning at 720M}: prompt-level strict accuracy, with greedy decoding and the Tulu chat template. Epoch 0 is the pretrained model.}
\label{fig:sft_ifeval}
\end{figure}

\paragraph{Instruction following.} Longer tuning does improve the instruction following performance. 
We evaluate IFEval~\citep{zhou2023ifeval}, all the runs exhibit similar performance and consistent and smooth improvements. The also demonstrates that the instruction-following ability is independent to its reasoning accuracy, even with the chat-template with aligned assistant instructions. 

\paragraph{Selection after tuning.} The lead of \method persists at $\lambda=16$ and $\lambda=64$. With plain prompts, At $\lambda=16$ it leads uniform baseline by 5.0 points before tuning and by 3.9 points after five epochs on the seven reasoning tasks; With chat template it leads by 3.7 points after five epochs. 
At $\lambda=64$ the lead keeps 2.4 points in plain format and 3.2 in chat format.

\begin{table}[ht!]
\centering
\caption{\textbf{Best accuracy over five instruction tuning epochs at 720M}. Every value is the highest over the five tuned models of a run on Tulu-v3. Zero-shot accuracy (\%) with plain prompts on each task reports \texttt{acc\_norm} of the five benchmarks, \texttt{acc} of WinoGrande and OpenBookQA; the mean takes the average over seven reasoning tasks with \textit{plain prompts} and with the \textit{Tulu chat template}. IFEval applies the prompt-level strict accuracy (greedy decoding, chat template). \method and similarity select with the FineWeb-Edu target.}
\vspace{0.3em}
\label{tab:sft_epochs}
\footnotesize
\setlength{\tabcolsep}{3.0pt}
\begin{tabular}{llcccccccccc}
\toprule
& & \multicolumn{5}{c}{Five benchmarks} & \multicolumn{2}{c}{Two other tasks} & \multicolumn{2}{c}{Mean (7 tasks)} & \\
\cmidrule(lr){3-7}\cmidrule(lr){8-9}\cmidrule(lr){10-11}
$\lambda$ & Scorer & ARC-E & ARC-C & PIQA & SciQ & HellaSw. & WG & OBQA & Plain & Chat & IFEval \\
\midrule
\multicolumn{2}{l}{Uniform} & 47.94 & 26.45 & 68.72 & 82.30 & 45.97 & 53.91 & 21.20 & 48.96 & 46.62 & 35.86 \\
\multicolumn{2}{l}{FineWeb-Edu} & 59.43 & 31.31 & 67.90 & 87.60 & 45.72 & 55.41 & 24.00 & 52.98 & 50.39 & 34.94 \\
\midrule
\multirow{2}{*}{4} & Similarity & \textbf{55.77} & \textbf{29.61} & \textbf{69.15} & 85.90 & 46.61 & 52.01 & \textbf{22.80} & \textbf{51.56} & 48.16 & \textbf{38.45} \\
 & \cellcolor{daisyrow}\method & \cellcolor{daisyrow}43.43 & \cellcolor{daisyrow}28.75 & \cellcolor{daisyrow}67.41 & \cellcolor{daisyrow}\textbf{86.30} & \cellcolor{daisyrow}\textbf{47.73} & \cellcolor{daisyrow}\textbf{52.96} & \cellcolor{daisyrow}21.80 & \cellcolor{daisyrow}49.56 & \cellcolor{daisyrow}\textbf{49.58} & \cellcolor{daisyrow}36.78 \\
\midrule
\multirow{2}{*}{16} & Similarity & \textbf{57.95} & \textbf{31.74} & 66.92 & \textbf{84.60} & 44.06 & 52.25 & 23.20 & 51.33 & 49.53 & 36.41 \\
 & \cellcolor{daisyrow}\method & \cellcolor{daisyrow}56.73 & \cellcolor{daisyrow}31.31 & \cellcolor{daisyrow}\textbf{69.97} & \cellcolor{daisyrow}82.40 & \cellcolor{daisyrow}\textbf{46.60} & \cellcolor{daisyrow}\textbf{54.38} & \cellcolor{daisyrow}\textbf{24.80} & \cellcolor{daisyrow}\textbf{51.86} & \cellcolor{daisyrow}\textbf{50.78} & \cellcolor{daisyrow}\textbf{37.34} \\
\midrule
\multirow{2}{*}{64} & Similarity & 57.49 & 31.91 & 64.47 & 85.60 & 39.57 & \textbf{54.22} & 23.00 & 50.54 & 49.29 & \textbf{38.08} \\
 & \cellcolor{daisyrow}\method & \cellcolor{daisyrow}\textbf{61.24} & \cellcolor{daisyrow}\textbf{32.85} & \cellcolor{daisyrow}\textbf{67.19} & \cellcolor{daisyrow}\textbf{87.00} & \cellcolor{daisyrow}\textbf{44.77} & \cellcolor{daisyrow}53.12 & \cellcolor{daisyrow}\textbf{25.60} & \cellcolor{daisyrow}\textbf{52.71} & \cellcolor{daisyrow}\textbf{51.19} & \cellcolor{daisyrow}36.78 \\
\bottomrule
\end{tabular}
\end{table}

\newpage
\clearpage
\section{Additional Details on Multilingual Pretraining}
\label{app:multilingual}
This appendix gives the details of the multilingual runs of Section~\ref{sec:multilingual}. The detailed configurations are included in \S\ref{app:sec:setup-multilingual}; additional results are provided in \S\ref{app:sec:results-multilingual}.

\subsection{Setup}
\label{app:sec:setup-multilingual}
\paragraph{Data Pool and Model.} The pool consists of six languages of FineWeb2 version 2.1.1~\citep{penedo2025fineweb2}: French (360.1M documents, 290.0B tokens), German (496.0M, 404.4B), Chinese (636.1M, 888.2B), Spanish (441.3M, 345.7B), Portuguese (199.7M, 148.4B) and Italian (239.0M, 197.9B), which contains 2.37B documents and 2.27T tokens in total. 
The model is the 124M configuration of \S~\ref{app:sec:setup} with tied embeddings over 200,064 vocabulary from the multilingual tokenizer, which include 238.6M total model parameters. The other training configurations including the schedule, optimizer and \method settings are the same as in the FineWeb (English) 124M runs. 

\paragraph{Benchmark target.} 
We adopt the six languages' subsets from the \texttt{dev} split from Global-MMLU~\citep{singh2024globalmmlu}\footnote{\url{https://huggingface.co/datasets/CohereLabs/Global-MMLU/viewer/am/dev}} and half of the test split from INCLUDE~\citep{romanou2025include}\footnote{\url{https://huggingface.co/datasets/CohereLabs/include-base-44}}, separating by the question stems. Therefore, the other half of test split for evaluation has no contamination or leakage.

\paragraph{HQ-FineWeb2 target and reference baseline.} We adopt the multilingual-transformed FineWeb-Edu classifier\footnote{\url{https://huggingface.co/epfml/FineWeb2-HQ-PlusPlus-Classifier}} to assign quality scores, ranging from 0-5. 
We curate HQ FineWeb2 as documents with scores at least 3.0, holding $1.56\%$ of documents (39.1B tokens). We split 256 documents to serve as target, which are strictly excluded from the held-out FineWeb2 set used for evlauation.
The \emph{Uniform on HQ FineWeb2} baseline is trained on the same set of 39.1B tokens with quality-scores at least 3.0, which serves as a strict baseline.

\subsection{Results}
\label{app:sec:results-multilingual}

\begin{figure*}[ht!]
\centering
\includegraphics[width=\textwidth]{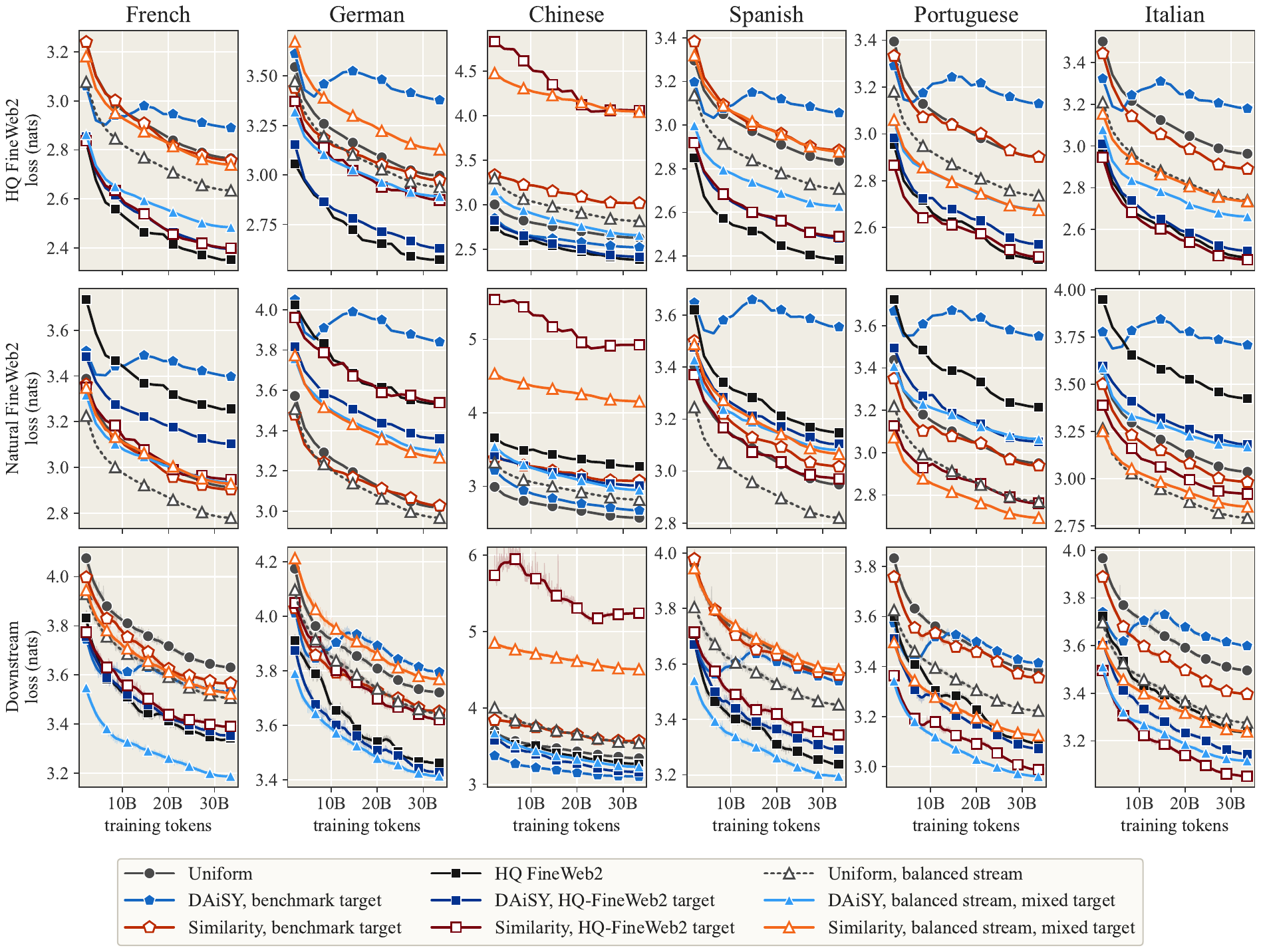}
\caption{\textbf{The runs of Figure~\ref{fig:multilingual} with the HQ-FineWeb2 and the benchmark target, per language}: held-out HQ FineWeb2 (top) and natural FineWeb2 loss (middle) at every permanent checkpoint, on the documents without a pool copy, and held-out benchmark loss (bottom; the mean of the language's Global-MMLU and INCLUDE tasks).}
\label{fig:multilingual_lang}
\end{figure*}

\paragraph{Losses per language.} Figure~\ref{fig:multilingual_lang} presents per-language losses. With the benchmark target, \method lowers the Chinese benchmark losses at every $\lambda$, by 0.12--0.15 nats on Global-MMLU and 0.18--0.35 nats on INCLUDE, whereas the losses on its German counterparts are above uniform across all candidate windows $\lambda=4$, $16$ and $64$. 
With HQ-FineWeb2 target and balanced stream with mixed target, compared with HQ FineWeb2 baseline, \method achieves improvements on downstream losses across all the five languages, except a slight worse on Spanish.
In comparison, \textit{Similarity} baseline always biased to Portuguese, while degrading on all the other languages.
\begin{table}[ht!]
\centering
\caption{\textbf{Multilingual online selection results at 124M.}}
\vspace{0.3em}
\label{tab:multilingual}
\footnotesize
\setlength{\tabcolsep}{4.5pt}
\begin{tabular}{llccccc}
\toprule
& & \multicolumn{3}{c}{Loss} & \multicolumn{2}{c}{Accuracy (\%)} \\
\cmidrule(lr){3-5}\cmidrule(lr){6-7}
$\lambda$ & Scorer & Down & HQ FW2 & Nat.\ FW2 & Global-MMLU & INCLUDE \\
\midrule
\multicolumn{7}{l}{\emph{Benchmark target, natural language mix}} \\
\multicolumn{2}{l}{Uniform (\textit{natural})} & 3.526 & 2.845 & 2.879 & 25.91 & 27.43 \\
\midrule
\multirow{2}{*}{4} & Similarity & 3.510 & \textbf{2.869} & \textbf{2.912} & 25.85 & 26.62 \\
 & \cellcolor{daisyrow}\method & \cellcolor{daisyrow}\textbf{3.447} & \cellcolor{daisyrow}2.872 & \cellcolor{daisyrow}3.151 & \cellcolor{daisyrow}\textbf{25.95} & \cellcolor{daisyrow}\textbf{28.23} \\
\midrule
\multirow{2}{*}{16} & Similarity & 3.517 & \textbf{2.905} & \textbf{2.998} & 25.92 & \textbf{27.79} \\
 & \cellcolor{daisyrow}\method & \cellcolor{daisyrow}\textbf{3.500} & \cellcolor{daisyrow}3.016 & \cellcolor{daisyrow}3.390 & \cellcolor{daisyrow}\textbf{26.18} & \cellcolor{daisyrow}27.57 \\
\midrule
\multirow{2}{*}{64} & Similarity & 3.572 & 2.990 & \textbf{3.133} & 25.77 & 27.43 \\
 & \cellcolor{daisyrow}\method & \cellcolor{daisyrow}\textbf{3.362} & \cellcolor{daisyrow}\textbf{2.930} & \cellcolor{daisyrow}3.375 & \cellcolor{daisyrow}\textbf{26.13} & \cellcolor{daisyrow}\textbf{29.83} \\
\midrule
\multicolumn{7}{l}{\emph{HQ-FineWeb2 target, natural language mix}} \\
\multicolumn{2}{l}{Uniform on HQ FineWeb2} & 3.274 & 2.434 & 3.305 & 27.31 & 29.76 \\
\midrule
\multirow{2}{*}{16} & Similarity & 3.607 & 2.807 & 3.473 & 26.27 & 28.15 \\
 & \cellcolor{daisyrow}\method & \cellcolor{daisyrow}\textbf{3.244} & \cellcolor{daisyrow}\textbf{2.489} & \cellcolor{daisyrow}\textbf{3.124} & \cellcolor{daisyrow}\textbf{26.54} & \cellcolor{daisyrow}\textbf{29.91} \\
\midrule
\multicolumn{7}{l}{\emph{Mixed target, balanced language stream}} \\
\multicolumn{2}{l}{Uniform (\textit{balanced})} & 3.443 & 2.761 & 2.824 & 25.89 & 28.81 \\
\midrule
\multirow{2}{*}{16} & Similarity & 3.626 & 3.046 & 3.242 & 25.74 & 27.06 \\
 & \cellcolor{daisyrow}\method & \cellcolor{daisyrow}\textbf{3.185} & \cellcolor{daisyrow}\textbf{2.664} & \cellcolor{daisyrow}\textbf{3.072} & \cellcolor{daisyrow}\textbf{26.57} & \cellcolor{daisyrow}\textbf{29.10} \\
\bottomrule
\end{tabular}
\end{table}

\begin{figure*}[ht!]
\centering
\includegraphics[width=\textwidth]{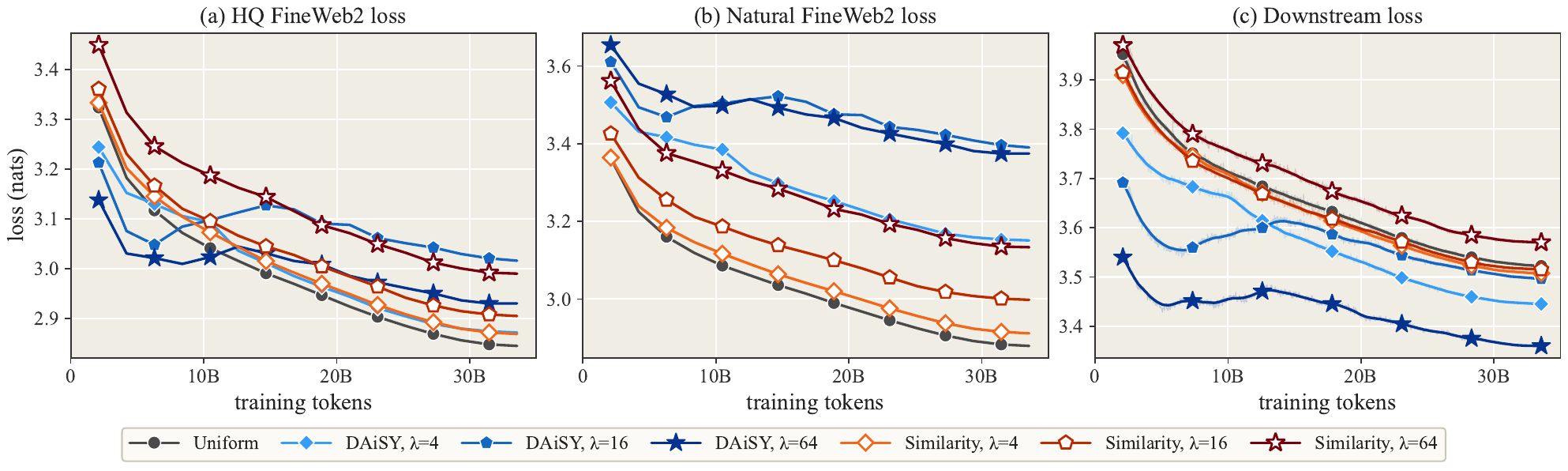}
\caption{\textbf{Multilingual runs with the benchmark target at $\lambda=4$, 16 and 64}.}
\label{fig:fw2_bench}
\end{figure*}

\begin{figure*}[ht!]
\centering
\includegraphics[width=0.67\textwidth]{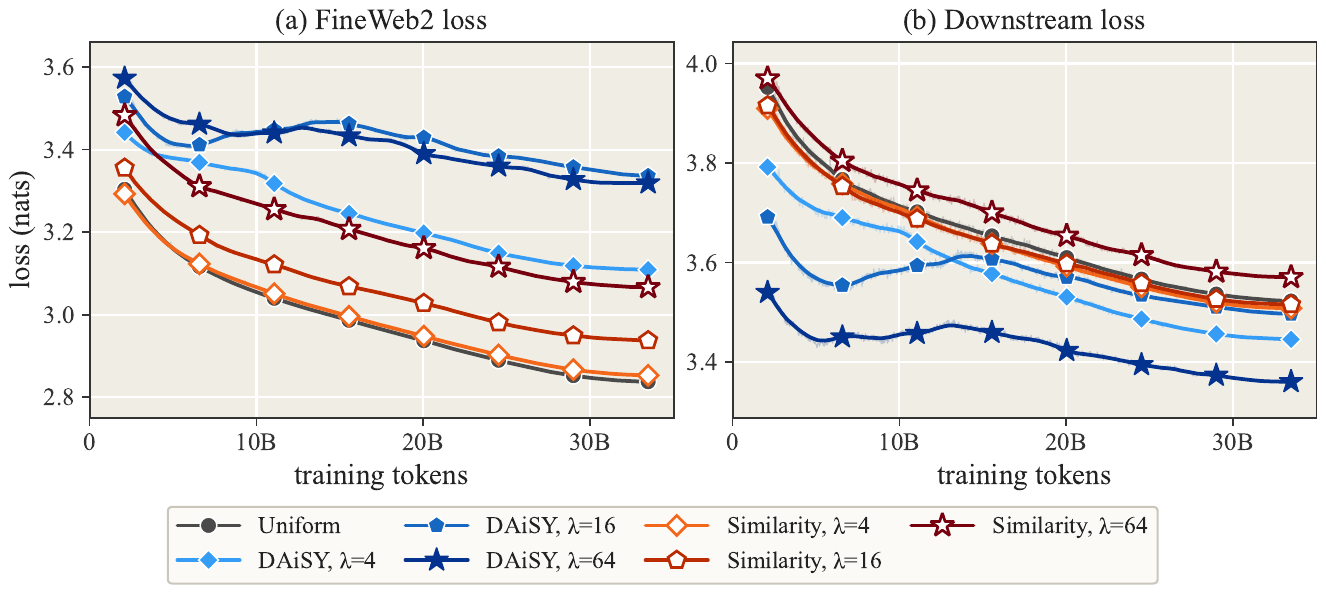}
\caption{\textbf{Multilingual runs with the benchmark target}. (a)~Held-out FineWeb2 loss and (b)~downstream loss, the average held-out loss on the 12 benchmark tasks (6 languages $\times$ 2 benchmarks).}
\label{fig:fw2_lambda}
\end{figure*}

\newpage
\clearpage
\section{Selection from a Token-Limited Pool}
\label{app:budget}
This appendix include the details for token-limited setup in  Section~\ref{sec:budget} over training.

\begin{figure*}[ht!]
\centering
\includegraphics[width=0.8\textwidth]{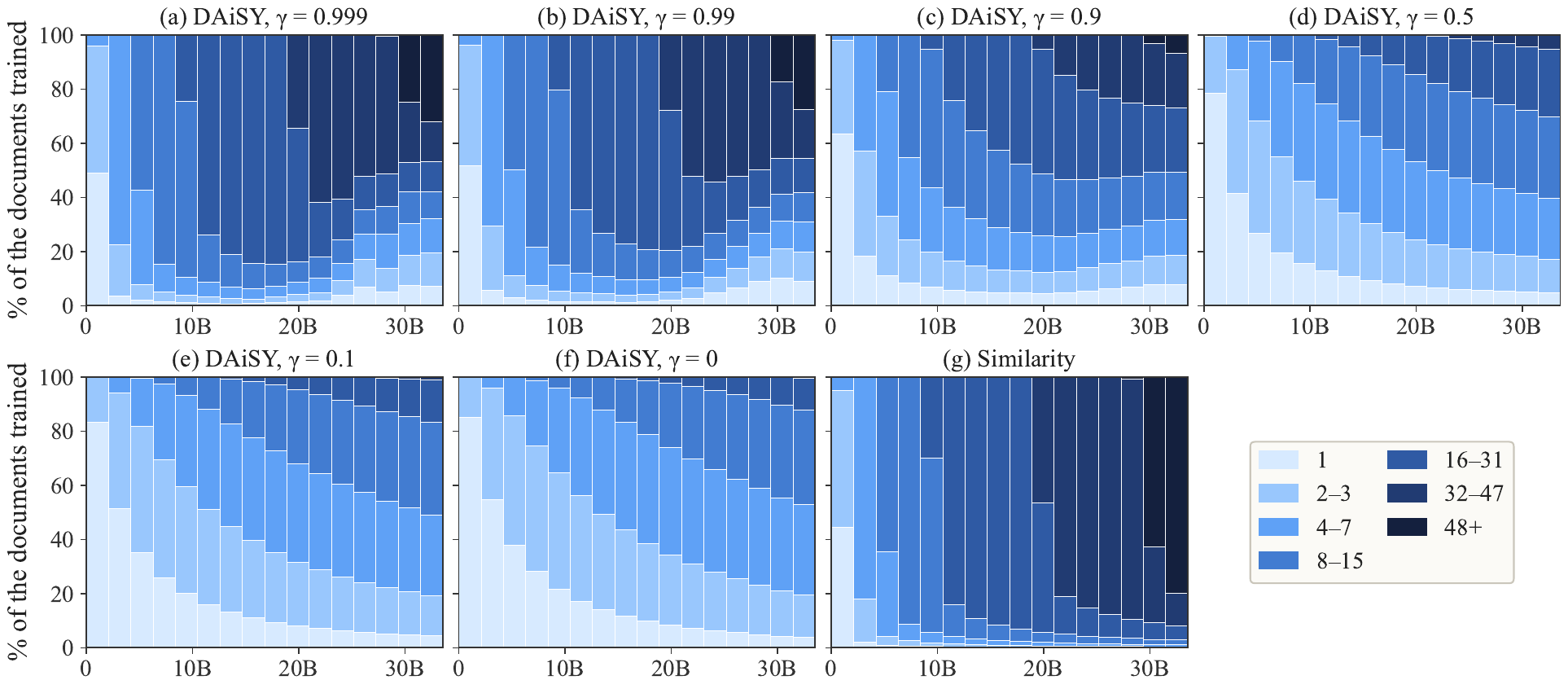}
\caption{\textbf{How often the rank-target runs train each document of the 10B subset across memory factor $\gamma$.}}
\label{fig:budget_rank}
\end{figure*}

\paragraph{Repetition.} We track the share of documents with various time of repetitions in each training batch along the run. Figure~\ref{fig:budget_rank} provides the repetitions in the 10B token-limited run under \textit{Similarity} and \method with various $\gamma$. \textit{Similarity} exhibits un-recoverable overfitting because of persistent repetition; while \method with $\gamma=0.999$ and $0.99$ turns to acquire more diversity after the severe overfitting and keeps a proper trade-off between task-relevance and diversity with $\gamma=0.9$.

\begin{figure*}[ht!]
\centering
\includegraphics[width=0.8\textwidth]{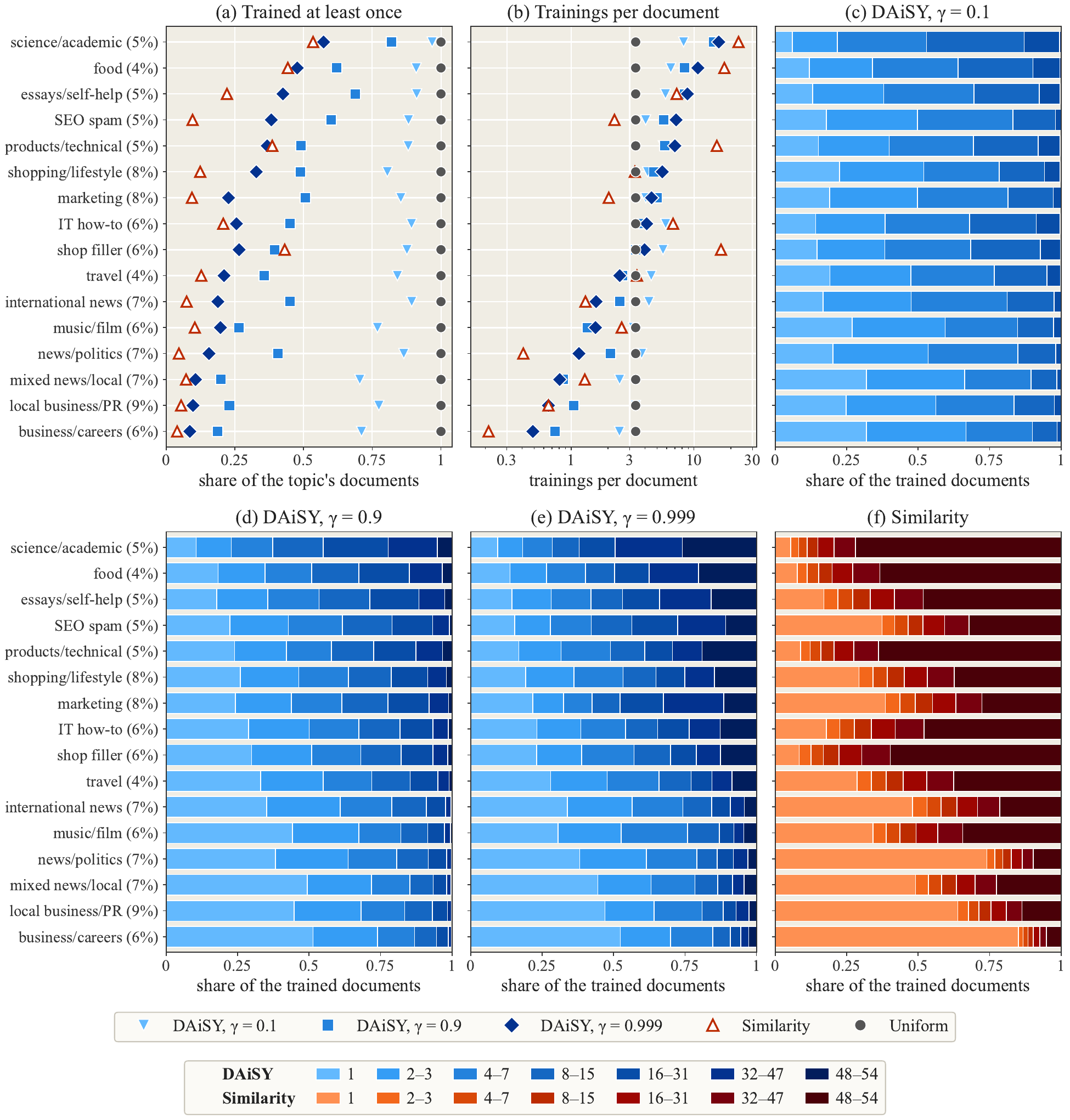}
\caption{\textbf{Repetition by topic on the 10B subset} at the end of training. \textbf{(a)}~Share of a topic's documents trained at least once. \textbf{(b)}~Trainings per document of the topic. \textbf{(c--f)}~The trained documents of each topic by the number of times they were trained.}
\label{fig:repeat_topics}
\end{figure*}

\paragraph{The selectors concentrate on the same topics, and a shorter memory spreads the selection.} 
Uniform training trains every document three or four times, in every topic (Figure~\ref{fig:repeat_topics}). 
\textit{Similarity} and \method at $\gamma=0.999$ mostly concentrated on the \textit{science/academic} topic with largest amount of repetitions; A shorter memory on \method reduces the concentration: at $\gamma=0.9$, \method as seen very few documents with $\geq48$ times repetitions, and covers $42.6\%$ of the subset and increases to $83.7\%$ at $\gamma=0.1$ (83.7\%).

\newpage
\clearpage
\subsection{Penalizing redundancy on the subset}
\label{app:redundancy}
Because it selects by utility alone, \method has no term for redundancy: near-duplicates receive nearly equal scores and can both be selected (\S\ref{sec:scorer}), and on the subset the same documents are selected in pass after pass (\S\ref{sec:budget}). Section~\ref{sec:budget} defines the four penalties that we add (Eqs.~\ref{eq:pen_count}--\ref{eq:pen_mom}), each as a single change to the rank-target run at $\gamma=0.9$ in Figure \ref{fig:budget_main}. We provide detailed results in Table~\ref{tab:redundancy}, Figure~\ref{fig:redundancy}. 

\begin{itemize}
\item \emph{Count penalty} ($\beta=0.5$). The selector penalizes the repetition by adding the log-scale of repeat time, universally applied on all documents.
\item \emph{Embedding penalty} ($\beta=15.41$). We set $\beta$ to allow the penalty scores has a standard deviation of 1, as performing a standardization. It is estimated before the run across 200 iterations with candidate window $\lambda=16$. 
\item \emph{Gradient penalty} and \emph{Momentum penalty}: we apply $\beta=1$. 
\end{itemize}

\begin{table}[ht!]
\centering
\caption{\textbf{Performance under redundancy penalties on the 10B subset at 124M.} \method ($\gamma=0.9$) serves as a reference without any penalty terms here.}
\label{tab:redundancy}
\footnotesize
\setlength{\tabcolsep}{4pt}
\begin{tabular}{lr rrr rrr}
\toprule
& & \multicolumn{3}{c}{Steps 120k} & \multicolumn{3}{c}{End of Training (256k)} \\
\cmidrule(lr){3-5}\cmidrule(lr){6-8}
Penalty & $\beta$ & Down & FineWeb & Used & Down & FineWeb & Used \\
\midrule
\method ($\gamma=0.9$) & -- & 3.345 & 3.214 & 29.6\% & 3.325 & 3.081 & 42.6\% \\
Count penalty & 0.5 & 3.420 & 3.070 & 58.0\% & 3.352 & 2.934 & 78.5\% \\
Embedding penalty & 15.41 & 3.412 & 3.107 & 44.2\% & 3.369 & 2.991 & 59.1\% \\
Gradient penalty & 1 & 3.349 & 3.171 & 36.9\% & 3.313 & 3.044 & 49.1\% \\
Momentum penalty & 1 & 3.342 & 3.161 & 37.4\% & 3.305 & 3.043 & 48.5\% \\
\midrule
\multicolumn{8}{l}{\emph{Similarity scorer}} \\
\method ($\gamma=0.9$) & -- & 3.385 & 3.391 & 16.3\% & 3.426 & 3.394 & 16.9\% \\
Count penalty & 0.5 & 3.362 & 3.255 & 37.6\% & 3.362 & 3.158 & 50.1\% \\
\bottomrule
\end{tabular}
\end{table}

\end{document}

%% file: math_commands.tex
\usepackage{amsmath,amsfonts,bm}

\def\eqref#1{equation~(\ref{#1})}

\def\1{\bm{1}}

\def\vzero{{\bm{0}}}

\def\vtheta{{\bm{\theta}}}
\def\va{{\bm{a}}}
\def\vb{{\bm{b}}}

\def\vm{{\bm{m}}}

\def\vp{{\bm{p}}}

\def\vs{{\bm{s}}}

\def\vu{{\bm{u}}}
\def\vv{{\bm{v}}}
\def\vw{{\bm{w}}}

\def\mA{{\bm{A}}}

\def\mG{{\bm{G}}}

\def\mI{{\bm{I}}}

\def\mP{{\bm{P}}}

\def\mV{{\bm{V}}}

\DeclareMathAlphabet{\mathsfit}{\encodingdefault}{\sfdefault}{m}{sl}
\SetMathAlphabet{\mathsfit}{bold}{\encodingdefault}{\sfdefault}{bx}{n}

\def\gB{{\mathcal{B}}}

\def\gL{{\mathcal{L}}}

\def\gS{{\mathcal{S}}}

\def\gV{{\mathcal{V}}}
\def\gW{{\mathcal{W}}}

\newcommand{\R}{\mathbb{R}}

\DeclareMathOperator*{\argmax}{arg\,max}


%% file: figures/pipeline_diagram.tex
\definecolor{selC}{HTML}{2E63A6}\definecolor{labC}{HTML}{BF5B17}\definecolor{fitC}{HTML}{008C7A}
\definecolor{inkC}{HTML}{1B2A33}\definecolor{mutC}{HTML}{636A6D}\definecolor{paperC}{HTML}{FBFAF6}
\begin{tikzpicture}[x=1cm,y=1cm,font=\footnotesize,line cap=round,
  lane/.style={rounded corners=7pt,draw=#1!45,fill=#1!7,line width=.55pt},
  head/.style={font=\fontsize{10}{12}\selectfont\bfseries,text=#1!85!black,anchor=west},
  sub/.style={font=\scriptsize\itshape,text=#1!80!black,anchor=west},
  card/.style={rounded corners=3.5pt,draw=#1!70,fill=white,line width=.6pt,align=center,inner sep=3.5pt,
               drop shadow={shadow xshift=.5pt,shadow yshift=-.6pt,opacity=.18}},
  chip/.style={rounded corners=5pt,fill=#1,text=white,font=\tiny\bfseries,inner xsep=3.2pt,inner ysep=1.7pt},
  arr/.style={-{Stealth[length=2.5mm,width=2mm]},line width=1pt,draw=#1},
  arr/.default=inkC,
  capt/.style={font=\scriptsize,align=center,text=inkC},
  ttl/.style={font=\fontsize{8}{9.5}\selectfont\bfseries,text=inkC,anchor=north west,inner sep=0pt}]
\newcommand{\docicon}[3]{%
  \begin{scope}[shift={(#1,#2)}]
    \path[draw=mutC!80,fill=#3,line width=.3pt,rounded corners=.6pt] (0,0) rectangle (.24,.31);
    \draw[white!55!mutC,line width=.3pt] (.05,.23)--(.19,.23) (.05,.155)--(.19,.155) (.05,.08)--(.15,.08);
  \end{scope}}
\newcommand{\circled}[2]{\tikz[baseline=(c.base)]\node[circle,fill=#1,text=white,inner sep=.7pt,font=\footnotesize\bfseries,minimum size=4.1mm](c){#2};}
\draw[lane=selC] (0,3.9) rectangle (16.5,7.0);
\draw[lane=fitC] (0,.35) rectangle (7.55,3.6);
\draw[lane=labC] (7.8,.35) rectangle (16.5,3.6);
\node[head=selC] at (.12,6.72) {\circled{selC}{1}\; Data Select};
\node[chip=inkC,font=\footnotesize\bfseries,inner xsep=6pt,inner ysep=2.6pt,anchor=east] at (16.3,7.0) {Overhead:\; +5.6\% FLOPs \,$\mid$\, +6\% Memory};
\node[head=labC] at (7.92,3.29) {\circled{labC}{2}\; LLM Train \& Utility Label};
\node[head=fitC] at (1.3,3.29) {\circled{fitC}{3}\; Learn Selector};
\begin{scope}[yshift=.22cm]
\foreach \i/\c in {0/mutC!14,1/mutC!22,2/mutC!32}{
  \begin{scope}[shift={(.62+.13*\i,5.13-.1*\i)},scale=2.0]\path[draw=mutC,fill=\c,line width=.35pt,rounded corners=.6pt] (0,0) rectangle (.24,.31);
  \draw[white!40!mutC,line width=.3pt] (.05,.23)--(.19,.23) (.05,.155)--(.19,.155) (.05,.08)--(.15,.08);\end{scope}}
\node[capt] at (1.13,4.42) {FineWeb stream\\[-1pt]1.77B documents};
\foreach \r in {0,1,2}\foreach \q in {0,...,4}{\docicon{3.0+.3*\q}{5.5-.39*\r}{mutC!12}}
\draw[selC!70,line width=.6pt,dashed,rounded corners=3pt] (2.92,4.66) rectangle (4.56,5.89);
\node[capt] at (3.74,4.28) {candidate window\\[-1pt]$\lambda\times$ the batch's tokens};
\node[cylinder,shape border rotate=90,aspect=.28,draw=selC!80,fill=selC!14,minimum height=1.05cm,minimum width=1.05cm,line width=.6pt] (idx) at (6.15,5.28) {$\vphi(x)$};
\node[capt] at (6.15,4.28) {frozen document\\[-1pt]embeddings};
\node[card=selC,minimum width=3.15cm,minimum height=1.95cm] (score) at (9.3,5.15) {};
\node at (9.3,5.75) {$s_t(x)=\vw_t^{\top}\big[\vphi(x);1\big]$};
\foreach \l [count=\i] in {1.72,1.38,1.05,.74,.5,.3}{
  \fill[selC!\the\numexpr 92-\i*11\relax,rounded corners=.5pt] (8.38,5.46-.135*\i) rectangle ++(\l,.092);}
\node[chip=selC] at (10.28,4.46) {$<\!10^{-6}$\% FLOPs};
\foreach \r in {0,1,2}\foreach \q in {0,...,4}{\docicon{12.0+.3*\q}{5.5-.39*\r}{mutC!9}}
\foreach \a/\b in {0/0,1/0,2/0,3/0,0/1,1/1,4/1,2/2,4/2}{\docicon{12.0+.3*\a}{5.5-.39*\b}{selC!62}}
\node[capt] at (12.74,4.28) {candidates ranked\\[-1pt]by utility scores};
\node[capt,font=\small,anchor=west,text height=1.6ex,text depth=.4ex] at (13.72,5.73) {\textcolor{selC}{$\blacksquare$} selected};
\node[capt,font=\small,anchor=west,text height=1.6ex,text depth=.4ex] at (13.72,5.35) {\textcolor{mutC!50}{$\blacksquare$} discarded};
\draw[arr] (1.95,5.28)--(2.85,5.28);
\draw[arr] (4.62,5.28)--(5.5,5.28);
\draw[arr] (6.75,5.28)--(7.65,5.28);
\draw[arr] (10.95,5.28)--(11.9,5.28);
\end{scope}
\node[card=labC,minimum width=2.85cm,minimum height=2.2cm] (train) at (14.92,1.71) {};
\foreach \i in {0,1,2,3}{
  \fill[inkC!\the\numexpr 28+\i*14\relax,rounded corners=1pt] (14.4,.98+.31*\i) rectangle ++(1.04,.2);}
\draw[arr] (14.2,.92)--(14.2,2.17);
\node[capt,rotate=90] at (13.92,1.54) {forward};
\draw[arr] (15.64,2.17)--(15.64,.92);
\node[capt,rotate=90] at (15.92,1.54) {backward};
\node[ttl] at (13.63,2.65) {Training step on $\bm{\gB}_t$};
\draw[arr=selC,line width=1.25pt,rounded corners=4pt] (13.48,5.095)--(15.98,5.095)--(15.98,2.85);
\node[fill=white,draw=selC!55,rounded corners=2pt,inner sep=1.6pt,font=\scriptsize,text=selC!80!black] at (15.98,3.75) {batch $\gB_t$};
\node[card=labC,minimum width=3.55cm,minimum height=1.06cm] (gate) at (11.25,2.39) {};
\node[ttl] at (9.62,2.79) {Label Gate};
\node at (11,2.26) {$y(x)=\big\langle\mV_t,\nabla\bar\ell(x)\big\rangle$};
\node[chip=labC] at (12.62,1.86) {$+5.0$\% FLOPs};
\draw[arr] (13.47,2.39)--(13.07,2.39);
\node[card=labC,minimum width=3.55cm,minimum height=.86cm] (tgt) at (11.25,.99) {};
\node[ttl] at (9.62,1.3) {Target direction};
\node at (10.75,.86) {$\mV_t=\mP_t\nabla\gL_\val(\vtheta_t)$};
\node[chip=labC] at (12.62,.56) {$+0.6$\% FLOPs};
\draw[arr] (10.6,1.44)--(10.6,1.84);
\draw[arr=labC,line width=1.25pt] (9.43,2.39)--(7.34,2.39);
\node[fill=white,draw=labC!55,rounded corners=2pt,inner sep=1.6pt,font=\scriptsize,text=labC!80!black] at (8.5,2.39) {labels $y$};
\node[card=fitC,minimum width=3.72cm,minimum height=2.32cm] (rank) at (5.44,1.76) {};
\node[font=\scriptsize] at (5.44,2.56) {$r_s(x)=\Phi^{-1}\!\Big(\dfrac{\mathrm{rank}_s(x)-\frac12}{n_s}\Big)$};
\newcommand{\minihist}[6]{%
  \foreach \c [count=\i from 0] in {#6}{
    \ifnum\c>0
      \fill[#4] ({#1+#2*\i},1.02) rectangle ++({#2*.86},{#3*\c});
      \ifnum\c<3 \fill[#4] ({#1+#2*(\i+.43)},{1.02+#3*\c+.05}) circle (.02);\fi
    \fi}
  \draw[inkC!55,line width=.35pt] ({#1-.03},1.02) -- ++({#2*#5+.06},0);}
\minihist{5.74}{.052}{.0225}{fitC}{28}{1,0,1,0,2,10,22,47,48,33,13,8,9,8,2,0,1,2,1,2,1,1,0,0,0,0,0,1}
\minihist{3.72}{.088}{.0327}{selC}{15}{1,2,5,9,17,24,32,33,32,24,17,9,5,2,1}
\draw[-{Stealth[length=1.6mm,width=1.3mm]},line width=.55pt,draw=inkC] (5.62,1.5)--(5.14,1.5);
\node[capt] at (6.47,.8) {raw labels $y$};
\node[capt] at (4.38,.8) {Gaussian ranks};
\node[card=fitC,minimum width=2.5cm,minimum height=2.16cm] (ridge) at (1.55,1.76) {};
\node[font=\fontsize{5}{6}\selectfont] at (1.55,2.48) {$\vw_t=\big(\mA_t+\bar\lambda_t\mI\big)^{-1}\vb_t$};
\foreach \a in {0,...,5}\foreach \b in {0,...,5}{
  \pgfmathsetmacro{\v}{int(8+60*exp(-abs(\a-\b)*.9))}
  \fill[fitC!\v] (.785+.19*\a,1.93-.19*\b) rectangle ++(.18,.18);}
\foreach \b in {0,...,5}{\pgfmathsetmacro{\v}{int(18+13*mod(\b*7,5))}\fill[inkC!\v] (2.125,1.93-.19*\b) rectangle ++(.18,.18);}
\node[chip=fitC] at (1.55,.68) {$<\!10^{-4}$\% FLOPs};
\draw[arr] (rank.west)--(ridge.east);
\draw[arr=fitC,line width=1.25pt,rounded corners=4pt] (1.1,2.84) -- (1.1,3.75) -- (9.3,3.75) -- (score.south);
\node[fill=white,draw=fitC!55,rounded corners=2pt,inner sep=1.6pt,font=\scriptsize,text=fitC!80!black] at (5.2,3.75) {updated selector weights $\vw_t$};
\end{tikzpicture}